\documentclass[a4paper,fleqn]{cas-dc}
\shorttitle{}                
\makeatletter
\let\ps@pprintTitle\ps@empty 
\renewcommand{\printorcid}[1]{} 
\makeatother
\usepackage{graphicx}
\usepackage{cite}
\usepackage[authoryear]{natbib}
\usepackage{amsmath,amsfonts}
\usepackage{mathrsfs}
\usepackage{algorithmic}
\usepackage[ruled,linesnumbered,noline]{algorithm2e}
\usepackage{array}
\usepackage{diagbox}
\usepackage{booktabs}
\usepackage{bm}
\usepackage{textcomp}
\usepackage{url}
\usepackage{verbatim}
\usepackage{multirow}
\usepackage{textcomp}
\usepackage{balance}
\usepackage{pifont}
\usepackage{color}
\usepackage{amsthm}
\usepackage{threeparttable}
\usepackage{caption}
\usepackage{subcaption}
\usepackage{graphicx}
\usepackage{float}
\definecolor{ccr}{RGB}{10,110,150}
\usepackage{hyperref}
\hypersetup{citecolor=ccr}

\definecolor{myblue}{rgb}{0.05,0.1,0.4}

\def\tsc#1{\csdef{#1}{\textsc{\lowercase{#1}}\xspace}}
\tsc{WGM}
\tsc{QE}

\usepackage{graphicx}
\begin{document}
\let\WriteBookmarks\relax
\def\floatpagepagefraction{1}
\def\textpagefraction{.001}
\let\printorcid\relax 



\title[mode = title]{Residual-Projected Multi-Collaboration Closed-Loop and Single Instance Quantization for Visually Impaired Assistance}

\tnotemark[1]


\author[1]{Xuanyu Wang}
\fnmark[1]
\ead{xywangmars@gmail.com} 

\author[2]{Haisen Su}
\fnmark[1]
\ead{suhaisen77@gmail.com} 

\author[5]{Jingtao Zhang}
\ead{jtzhanguestc@gmail.com}

\author[1,3,6]{Xiangxiang Wang}
\fnmark[$\ast$]
\ead{xxwang@uestc.edu.cn}

\author[1]{Yongbin Yu}
\ead{ybyu@uestc.edu.cn}

\author[1]{Manping Fan}
\ead{fmpfmp@uestc.edu.cn} 

\author[3]{Jialing Xiao}
\ead{jollyxiao1998@163.com} 

\author[3,4]{Bo Gong}
\ead{gongbo@med.uestc.edu.cn}

\author[2]{Siqi Chen}
\ead{chensq0503@163.com} 

\author[1]{Mingsheng Cao}
\ead{cms@uestc.edu.cn} 

\author[1]{Liyong Ren}
\ead{lyren@uestc.edu.cn} 

\author[3,5]{Zhenglin Yang}
\fnmark[$\ast$]
\ead{yangzhenglin@cashq.ac.cn}

\address[1]{School of Information and Software Engineering, University of Electronic Science and Technology of China, Chengdu, Sichuan, China}

\address[2]{School of Mathematical Sciences, Sichuan Normal University, Chengdu, Sichuan, China}

\address[3]{Sichuan Provincial Key Laboratory for Human Disease Gene Study, Sichuan Academy of Medical Sciences \& Sichuan Provincial People's Hospital, 
	University of Electronic Science and Technology of China, Chengdu, Sichuan, China}

\address[4]{Research Unit for Blindness Prevention, Chinese Academy of Medical Sciences, Sichuan Academy of Medical Sciences and Sichuan Provincial People’s Hospital, Chengdu, Sichuan, China}

\address[5]{School of Medicine, University of Electronic Science and Technology of China, Chengdu, Sichuan, China}

\address[6]{Tibetan Language Intelligence National Key Laboratory, Qinghai Normal University, Xining, Qinghai, China}

\fntext[1]{These authors contributed equally to this work.}

\cortext[1]{Corresponding authors}

\tnotetext[1]{This work was supported in part by the Fundamental Research Funds for the Central Universities under Grants ZYGX2025YGLH001 and Grant ZYGX2024XJ038, in part by the National Natural Science Foundation of China under Grants 62276055 and 62406062, in part by the Sichuan Science and Technology Program under Grants 2024NSFSC1476 and 2023YFG0288, in part by the Sichuan Provincial Major Science and Technology Project under Grant 2024ZDZX0012.}

\begin{abstract}
Visually impaired users face significant challenges in daily information access and real-time environmental perception, and there is an urgent need for intelligent assistive systems with accurate recognition capabilities. Although large-scale models provide effective solutions for perception and reasoning, their practical deployment on assistive devices is severely constrained by excessive memory consumption and high inference costs. Moreover, existing quantization strategies often ignore inter-block error accumulation, leading to degraded model stability. To address these challenges, this study proposes a novel quantization framework---Residual-Projected Multi-Collaboration Closed-Loop and Single Instance Quantization(RPIQ), whose quantization process adopts a multi-collaborative closed-loop compensation scheme based on Single Instance Calibration and Gauss-Seidel Iterative Quantization. Experiments on various types of large-scale models, including language models such as OPT, Qwen, and LLaMA, as well as vision-language models such as CogVLM2, demonstrate that RPIQ can compress models to 4-bit representation while significantly reducing peak memory consumption (approximately 60\%-75\% reduction compared to original full-precision models). The method maintains performance highly close to full-precision models across multiple language and visual tasks, and exhibits excellent recognition and reasoning capabilities in key applications such as text understanding and visual question answering in complex scenarios. While verifying the effectiveness of RPIQ for deployment in real assistive systems, this study also advances the computational efficiency and reliability of large models, enabling them to provide visually impaired users with the required information accurately and rapidly.
\end{abstract}
\maketitle
\section{Introduction}
Visually impaired individuals face significant challenges in daily information access\cite{abidi2024comprehensive}, including difficulties in efficiently retrieving textual content, gaining timely awareness of their surrounding environment, and obtaining precise answers when encountering obstacles. Traditional assistive tools mostly offer fixed modules and limited query capabilities, making them ill-suited to meet modern practical demands. Therefore, building an intelligent assistive system\cite{souza2024intelligent} capable of accurately understanding information, efficiently processing complex textual data, and providing reasonable suggestions based on user needs is of great significance for improving the quality of life for the visually impaired.

In recent years, large models have demonstrated immense potential in assistive tasks for the visually impaired\cite{zhao2024vialm}. Leveraging their powerful natural language processing capabilities and advanced features such as intelligent interaction and virtual assistance, these models are increasingly applied to build intelligent assistive systems\cite{li2025vocot}. Examples include Q\&A assistants, text interpretation tools, and environmental information query agents, providing visually impaired users with relevant information regarding access and interaction. However, such models typically contain billions or even trillions of parameters. Each inference request requires forward propagation through the entire model, leading to two core challenges\cite{hadi2023large,li2024efficient}: excessive memory consumption and skyrocketing inference costs. These issues have become critical bottlenecks hindering the large-scale deployment of these models and represent urgent problems to be solved in the field of large models.

To further advance large model development, Post-Training Quantization (PTQ) has emerged as a mainstream technical solution. This strategy achieves efficient quantization without retraining, significantly reducing inference overhead. Compared to Quantization-Aware Training (QAT), which requires massive training datasets, PTQ demonstrates significant advantages: it bypasses the data training process, adapts to various model quantization strategies, and has become the preferred solution for lightweight model optimization in practice\cite{dettmers20218}. Among existing PTQ methods, GPTQ is a representative one-shot quantization strategy based on second-order Hessian matrix information, achieving good accuracy on large models\cite{zhou2024lidar}.

To overcome the limitation of GPTQ algorithms where optimal solutions become trapped in local optima due to their greedy nature, single iteration characteristics\cite{frantar2022gptq}, and unidirectional error accumulation, a novel quantization scheme is proposed— Residual-Projected Multi-Collaboration Closed-Loop and Single instance Quantization for Visually Impaired Assistance (RPIQ). This approach abandons GPTQ's static logic of single-pass blockwise greedy optimization, establishing a multi-round iterative residual correction framework centered on the core objective of minimizing output spatial error. Through a two-stage process of initial quantization generation followed by iterative correction of output residuals, it achieves higher accuracy compared to GPTQ. 

Specifically, the RPIQ quantization scheme proposed in this paper adopts the same quantization method as GPTQ in the first stage, ensuring the global validity of statistical information regarding model weight importance. Through forward propagation of input data, it generates activation information for the corresponding layer, providing a complete basis for Hessian matrix computation. By replicating GPTQ's second-order logic optimization, the Hessian matrix computed from global activation data in RPIQ's first stage precisely quantifies each weight's contribution to the model's output loss, guaranteeing the global rationality of weight optimization. The first stage retains GPTQ's unidirectional, one-time blockwise greedy quantization approach. It divides the weight matrix into blocks of fixed size and rapidly obtains locally optimal weights for each block via Cholesky decomposition during optimization. Following Phase 1, the RPIQ framework retains critical information including the global Hessian matrix and initial quantized weights in memory rather than storing only the final quantized weights. This retained data serves as the foundation for Phase 2 iterations, eliminating redundant computation of global information.

The second phase of RPIQ focuses on addressing the issue of accumulated block-to-block errors caused by unidirectional, one-time block-by-block optimization in GPTQ. In this phase, there is no need for additional calibration data. Instead, the final batch of data from the global calibration process is directly invoked in memory. By directly reading the input matrix $X_{last}$ corresponding to this batch of data and the original output matrix $Y_{origlast}$ . This design effectively avoids the time overhead of repeatedly reloading calibration data, thereby improving the overall execution efficiency of the algorithm. The second stage adopts a multi-round block-level cooperative iterative correction strategy. In each iteration, based on the input matrix \(X_{last}\) corresponding to the in-memory calibration data and the global Hessian matrix, the method dynamically computes the difference between the original and quantized outputs for each block, \(D = Y_{origlast} - Y_{quaninit}\), to obtain the corresponding global output residual that accurately reflects the current quantization error. Meanwhile, the second stage employs a Gauss-Seidel-type iterative update scheme, where the optimization of the current block is immediately performed using the latest weights of the already updated blocks, effectively preventing GPTQ blocks from being optimized in an isolated or mutually independent manner.

Based on known information, there is currently a scarcity of systematic research capable of maintaining the robust performance of GPTQ while compensating for inter-block errors through a multi-round blockwise collaborative iterative refinement scheme, all without relying on any external activation information during the residual correction process. For instance, GPTQT employs a two-stage progressive quantization procedure\cite{guo2024gptqt}: it first performs linear quantization to obtain an initial draft, and subsequently implements a second-stage quantization by optimizing binary codes and utilizing residual information to minimize output error. This indicates that the overall methodology remains in an early stage of exploration, highlighting an urgent need to address this issue and enrich the relevant body of research.

Inspired by the aforementioned discussion, this paper focuses on the block-level bias introduced by the greedy one-shot optimization paradigm within PTQ frameworks, as well as its impact on final quantization accuracy. The main contributions of this study are summarized as follows:

\vspace*{-5pt}
\begin{itemize}
\setlength{\itemsep}{0pt}
\setlength{\parsep}{0pt}
\setlength{\parskip}{0pt}
\item[(1)]
	Block based multi-collaborative closed-loop compensation structure based on residual adjustment is designed, which by explicitly constructing global residuals and blockwise directional residuals during the quantization process, effectively mitigates the one-way error accumulation caused by one-shot greedy quantization and addresses the issue of poor stability of large models in visual assistance tasks.
    
\item[(2)]
    Single instance calibration paradigm is designed based on an instantaneous Hessian curvature reconstruction scheme, which innovatively introduces a single batch of calibration data as the input source. Under this framework, the second-stage iterations only need to directly access the last batch of data stored in memory from the global calibration process, effectively avoiding the time overhead of repeatedly reloading calibration data, improving the execution efficiency of the algorithm, and making it more suitable for deployment on resource-constrained assistive devices for visually impaired users.

\item[(3)]
     Gauss-Seidel governed dynamic blockwise iterative quantization scheme is designed, which when quantizing the $i$ block, directly utilizes the latest weights of blocks $1$ to $i-1$ that have been optimized in the current iteration, thereby achieving a block-level iterative quantization collaborative compensation mechanism that provides a more robust strategy for deploying large models in visually impaired assistance application scenarios.
\end{itemize}

The rest of this paper is organized as follows. Section \ref{sec:related} introduces related work, covering the application of large models in assistive systems for visually impaired users, the development of post-training quantization methods, and the limitations of existing approaches in calibration data dependency and iterative optimization. Section \ref{sec:main_algorithm} elaborates on the three core designs of the proposed RPIQ quantization strategy, including block based multi-collaborative closed-loop compensation structure based on residual, single instance calibration paradigm, and Gauss-Seidel governed dynamic blockwise iterative quantization. Section \ref{sec:Experiment} evaluates the performance of various language models and vision-language models through experiments, analyzing task performance and resource consumption. Section \ref{sec:Discussion} discusses the research results, insights, methodological advantages and limitations. Finally, Section \ref{sec:Conclusion} concludes the paper and outlines directions for future research.


\section{Related Work}\label{sec:related}
With the rapid development of large models\cite{hadi2023large,li2024efficient} and the continuous advancement in their text understanding capabilities in recent years, an increasing number of studies are applying them to assistive systems for visually impaired users to enhance tasks such as text reading comprehension and question answering. In these application scenarios, assistive systems are often required to run continuously over long periods under limited hardware conditions, while demanding high accuracy and stability in their semantic responses and explanations. This makes the deployment of highly accurate and robust quantized large models under strict memory and inference budgets a critical issue.

\subsection{Assistive Systems for the Visually Impaired and Large Models}
In recent years, a growing body of research has emerged concerning information assistance for the visually impaired, with intelligent assistive systems driven by large models\cite{xie2025beyond} becoming the mainstream technology. This technology effectively avoids the drawbacks of earlier systems that relied heavily on fixed models, while offering intelligent responses and more reasonable explanations in complex environments.

As large models have shown significant improvements in language understanding and question-answering capabilities, several studies have begun to apply them to visual assistance tasks\cite{baig2024ai}. Holiel et al. constructed a large model-based conversational assistance system that provides visually impaired users with information retrieval and answers to daily life questions through natural language interaction\cite{holiel2024assisting} . Similarly, Leporini et al. investigated the practical performance of several mainstream generative tools among the blind population, highlighting the huge advantages of large models in Q\&A and explanation, but also exposing the issue of occasional incorrect feedback\cite{leporini2025preliminary}.  Furthermore, as the scale of parameters and inference costs continue to rise during the quantization process, directly deploying large models in practical systems makes long-term stable operation difficult in resource-limited environments. Based on the issues mentioned above, the RPIQ quantization algorithm proposed in this paper aims to improve accuracy and explanatory capability by refining the quantization algorithm, allowing large models to be better and more widely applied in visual assistance tasks.

\subsection{Development of Model Quantization Methods}
In recent years, driven by the widespread application of large models and Deep Neural Networks (DNNs), model quantization techniques have garnered extensive research attention\cite{gholami2022survey,nagel2106white}. Current mainstream quantization techniques are primarily categorized into PTQ and QAT. QAT approaches optimize weight robustness against quantization perturbations by introducing simulated quantization and Straight-Through Estimators (STE) during the fine-tuning stage, thereby achieving exceptional accuracy retention. However, the computational intensity of QAT renders it difficult to scale to LLMs characterized by massive parameter counts\cite{esser2019learned}. Consequently, PTQ has emerged as the prevailing choice for LLM quantization, owing to its lightweight nature that eliminates the need for retraining and relies on only a small amount of calibration data\cite{li2021brecq,nagel2020up}. Notably, in recent years, PTQ methods have achieved breakthrough advancements in mitigating accuracy loss. Among these, GPTQ stands out as a prominent representative. Its core innovation lies in formulating the quantization problem as a local optimization task and employing a column-wise greedy search mechanism to minimize the output reconstruction error of each layer\cite{frantar2022gptq}. While this scheme effectively reduces model quantization error and maintains hardware friendliness, its block-by-block optimization process lacks a comprehensive consideration of inter-block dependencies, leading to the accumulation of errors across blocks. Similarly, Loss Aware Post-training Quantization (LAPQ) proposed reducing quantization errors by jointly optimizing the quantization step sizes across all network layers\cite{nahshan2021loss}; however, it fails to address the issue of error accumulation between blocks during this joint optimization. The Activation-Aware Weight Quantization For On-Device LLM Compression And Acceleration (AWQ) approach utilizes activation information to identify and protect salient weights during quantization to minimize errors\cite{lin2024awq}. However, while it enhances accuracy through activation-aware quantization, it also fails to resolve the accumulation of inter-block errors during the quantization process. To address this, a multi-round block-level cooperative optimization strategy is designed—RPIQ, which realizes inter-block cooperative optimization and effectively resolves the issue of unidirectional error accumulation inherent in GPTQ.

\subsection{The Dependence of Quantitative Models on Calibration Data}
Nowadays, achieving the balance between quantization accuracy and calibration during the quantization process largely depends on the efficient utilization of the provided calibration data \cite{gholami2022survey}. For assistive systems targeting visually impaired users, the actual quantization process is often constrained by the limited resources of the server, making it difficult to repeatedly load large-scale calibration data. This makes multi-round traversal schemes, traditionally reliant on full calibration data, challenging to apply. Among various model compression techniques, PTQ has become the dominant quantization strategy for large models due to its significant advantage of not requiring model retraining \cite{nagel2019data,yao2022zeroquant}. Within the realm of PTQ methods, Adaptive Rounding (AdaRound) and Block Reconstruction-based Post-Training Quantization (BRECQ) were early quantization solutions \cite{li2021brecq,nahshan2021loss}. They employ a strategy based on full data calibration, achieving significant accuracy improvements through multiple optimization passes over the entire calibration dataset. However, as the dataset and model scales increase, this approach incurs massive computational overhead. Especially when quantizing large-scale models, it frequently triggers Out-of-Memory (OOM) errors \cite{lin2024awq}.

Unlike these methods, GPTQ \cite{frantar2022gptq} adopts a quantization strategy based on a small-scale dataset, adjusting quantization via the second-order information of the Hessian matrix and greedy search. However, it simultaneously neglects the issue of inter-block error accumulation caused by its greedy nature. Similarly, AWQ \cite{lin2024awq} also improves quantization accuracy by activating on a small-scale dataset and selectively protecting critical weights. OMNI \cite{shao2023omniquant} employs a mixed-precision quantization strategy on a small-scale dataset, dynamically adjusting the quantization precision across different layers to enhance accuracy. However, neither of these two schemes considers the issue of quantization reversibility.

To balance quantization accuracy and efficiency, this paper proposes a novel quantization strategy: the Single instance calibration paradigm based on Instantaneous Hessian Curvature Reconstruction paradigm. Unlike traditional quantization schemes that require repeated use of the entire calibration dataset, this scheme, after constructing the global Hessian matrix, directly reads the resident last batch of calibration data from memory to guide the iterative residuals for the instance. This strategy retains the global second-order information while avoiding the dependence on the full dataset and the additional memory overhead during the optimization iteration phase. By retaining the global second-order information and avoiding reliance on the full dataset during the optimization iteration phase, this strategy is more suitable for completing the quantization process in resource-limited environments, providing a feasible solution for the subsequent deployment of large models in practical assistance tasks.
\subsection{Dilemma of iterative optimization}
In recent years, many works have attempted to further reduce quantization error by adopting iterative optimization methods\cite{wang2019haq,hubara2020improving,yao2022rapq,hubara2021accurate}. For instance, A Backpropagation-Free Algorithm for
Post-Training Quantization (COMQ) proposed to minimize quantization error through coordinate descent-style per-coordinate updates and multiple rounds of iteration within each layer for quantized weights and scaling factors\cite{zhang2025comq}. Similarly, Efficient Finetuning-Free Quantization for Asymmetric Calibration (GPTAQ) iteratively solves for the optimal quantization value of each weight element within the framework of Optimal Brain (OBC), achieving a higher-precision solution\cite{li2025gptaq} . Although these two schemes are more meticulous in local search, they usually start from uncalibrated initialization, which results in the need for many rounds of iteration to converge to a satisfactory solution. In contrast, the RPIQ proposed in this paper regards GPTQ as a high-quality approximate solution in the first stage and then implements fine-tuning near a good initial value to achieve rapid convergence with fewer iterations, thereby facilitating the application of large models in visual assistance scenarios that require high stability.

Regarding the iteration mechanism itself, existing iterative methods predominantly employ a coordinate-wise or element-wise update mechanism, or sequentially perform independent updates of weight values in each iteration. These are fundamentally local independent compensation mechanisms, making it difficult to effectively reduce inter-block errors. Building on this, this paper adopts the Gauss-Seidel iterative method to implement an iterative update guided by residuals mechanism. By constructing a directional residual and local least squares, it progressively eliminates the global output residual, thereby achieving multi-collaborative closed-loop compensation based on residua within a limited number of steps\cite{li2025gptaq,zhang2025comq}.


\section{Methods}\label{sec:main_algorithm}
In intelligent assistive systems for visually impaired users, large models typically function to understand user input and provide reasonable and accurate answers to user needs. Since it is difficult to deploy large models directly on actual devices, quantization techniques for large models have become a primary research direction. Quantization aims to reduce the parameter scale and computational and memory overhead of large models during inference by mapping floating-point weights to low-bit integers\cite{dettmers20218,lin2024awq}. In this section, RPIQ is systematically introduced from three aspects. First, a block based collaborative closed-loop compensation mechanism is constructed under the guidance of output residuals to explicitly reduce the accumulation of errors between blocks. Secondly, a single instance calibration paradigm based on instantaneous Hessian curvature reconstruction is proposed, which reduces the dependence on global calibration data while retaining the global second-order information. Finally, the Gauss-Seidel updated dynamic block iterative quantization mechanism is introduced, so that the quantization process can converge quickly and stabilize the value.

\subsection{Block based multi-collaborative closed-loop compensation mechanism based on output residuals}

\begin{figure*}[htbp]
    \centering
   \includegraphics[width=0.97\textwidth,trim=0.1cm 0.1cm 0.1cm 0.1cm,clip]{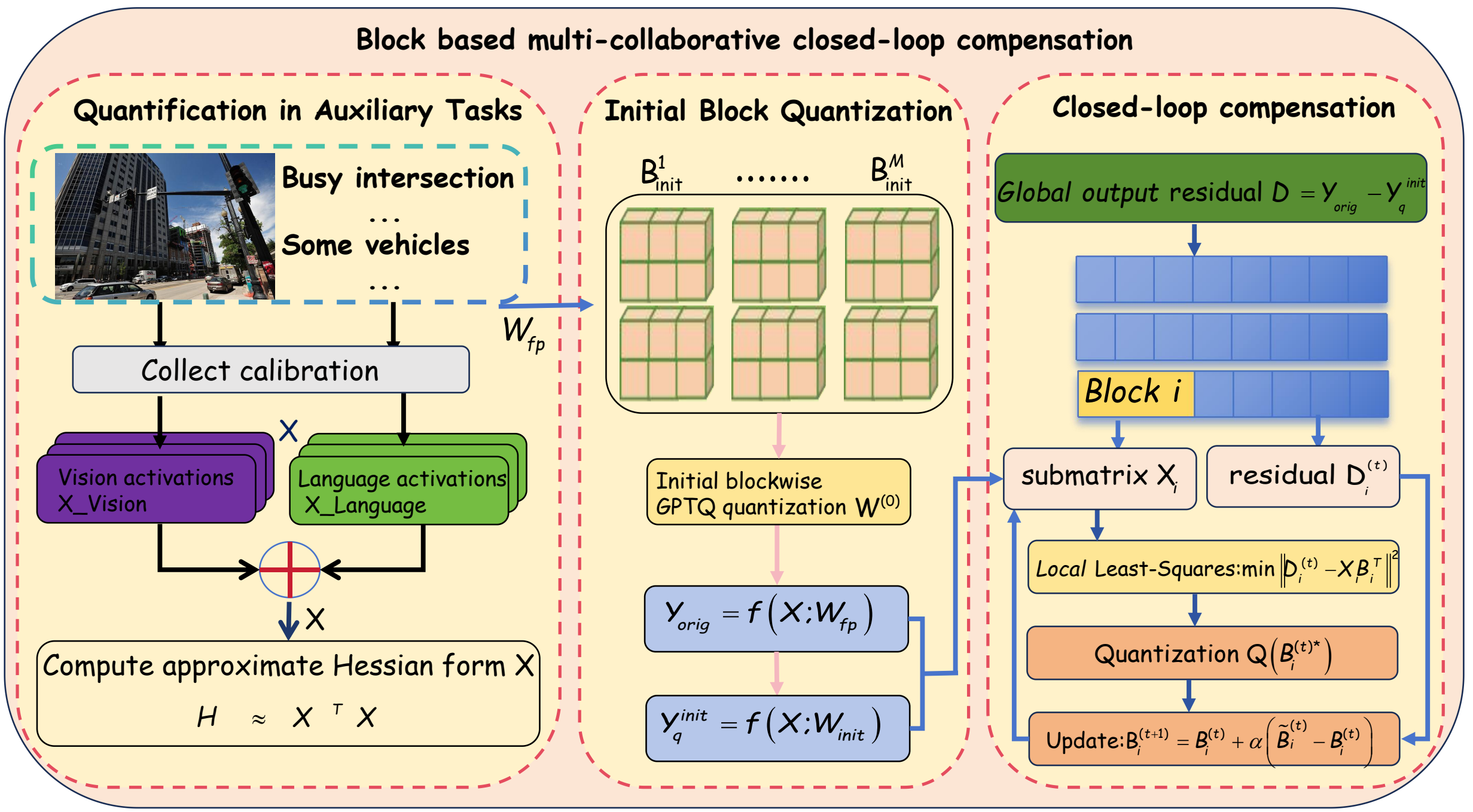}
   \caption{Block based multi-collaborative closed-loop compensation.}
   \label{fig:Block-level-coll}
\end{figure*}

In visually impaired assistance tasks, the output accuracy and stability of the quantized model directly influence the user's information acquisition accuracy. Traditional blockwise greedy quantization neglects the inter-block error accumulation problem, leading to significant performance degradation in tasks such as long-text reading comprehension and multi-turn question answering.

To address this, this paper introduces block based multi-collaborative closed-loop compensation structure based on residual adjustment, achieves multi-round inter-block cooperative mechanisms and alleviates the error accumulation problem. Specifically speaking, in the first stage of RPIQ quantization, the quantization strategy of GPTQ is completely followed, and the weight matrix of each layer is divided into several blocks by column $W \in \mathbb{R}^{C_{\text{out}}\times C_{\text{in}}}$ into several blocks by column ${B^{\text{init}}_{1}, B^{\text{init}}_{2}, \ldots, B^{\text{init }}_{M}}$,
where $C_{\text{out}}$ represents the number of output channels, $C_{\text{in}}$ represents the number of input channels, and $M$ represents the total number of blocks divided,
For each $B^{\text{init}}_{i}$ is the initial quantized weight matrix for the $i$ block. Given the full-precision input matrix $X \in\mathbb{R}^{N\times C_{\text{in}}}$, and the Hessian matrix information,
The initial quantization weight of each block can be solved, so as to obtain the initial quantization matrix
$W^{{init}} = [B^{{init}}_{1}, B^{{init}}_{2}, \dots, B^{{init}}_{M}]$. Using the same batch of input information $W_X$, the corresponding output $Y_{{orig}} = f(X; W_{fp})$, and $W^{{init}}$ obtained by the initial quantization weight $Y^{\text{init}}_{q} = f(X; W^{\text{init}})$, thus forming a full-precision quantization branch and a initial quantization branch in the graph. Where $C_{out}$ represents the number of output channels, $C_{in}$ represents the number of input channels, $M$ represents the total number of blocks divided, and for each $B_i^{init}$ is the initial quantization weight matrix of the $i$ block.

Given the full-precision input matrix $X \in \mathbb{R}^{N \times C_{in}}$, and the Hessian matrix information, the initial quantization weight of each block can be solved, so as to obtain the initial quantization $W^{init} = \left[ B_1^{init}, B_2^{init}, \dots, B_M^{init} \right]$. After replacing the full-precision weight $W_{fp}$ with $W^{init}$, the calibration input $X$ is propagated forward to obtain the corresponding initial quantization output:
\begin{equation}
Y_q^{init} = f(X; W^{init}), Y_{orig} = f(X; W_{fp}) 
\end{equation}
where $N$ is the number of samples, $C_{in}$ is the dimension of the input feature, $Y_q^{init}$ is the initial quantized output value, $Y_{orig}$ is the full-precision output value, and $f(X; W)$ represents the forward function calculated using input $X$ and the weight matrix $W$. Based on these two output branches, the definition of the global output is given:
\begin{equation}
    D = Y_{\text{orig}} - Y^{\text{init}}_{q}
\end{equation}

Thus, the global output residual can be measured, and further refine this residual into directional block-level errors $D_i^{(t)}$ in subsequent block-level collaborative compensation.

In order to reduce the problem of block-to-block error accumulation caused by one-way, one-time greedy calculations in GPTQ, the second stage of RPIQ is to compensate for errors around $D$. Note that the input submatrix corresponding to the $i$ block is $X_i$, and the quantization weight of the $j$ block in the current iteration is $B_j^{(t)}$, then the quantization output can be decomposed by block:
\begin{equation}
    Y_q^{(t)} = \sum_{j=1}^M Y_{q,j}^{(t)}, \quad Y_{q,j}^{(t)} = X_j (B_j^{(t)})^T
\end{equation}
where $Y_q^{(t)}$ represents the overall output of the current quantization network on input $X$ in the second stage $t$ iteration, and $Y_{q,j}^{(t)}$ represents the output part of the $j$ block contributed separately in the second stage $t$ iteration.

When iterating over the $t$ round and updating the $i$ block, RPIQ uses the current network state to construct the directed output residual for that block
\begin{equation}\label{eq:d}
    D_i^{(t)} = Y_{\text{orig}} - \left(Y_q^{(t)} - Y_{q,i}^{(t)}\right) 
\end{equation}

That is, the old contribution of the current block value is removed from the global residual, so that $D_i^{(t)}$ more accurately reflects the error to be calibrated when adjusting the $i$ block. On this basis, a local least squares problem is established for the $i$ block:
\begin{equation}\label{eq:target}
    \min_{B_i} \left|\left| D_i^{(t)} - X_i B_i^T \right|\right|_2^2 
\end{equation}

Thus, the problem is transformed into a given input matrix $X_i$, using the weight $B_i$ to fit the target $D_i^{(t)}$, and by using the Frobenius norm to expand and derive $B_i$ and make the derivative 0, the corresponding analytic solution: 
\begin{equation}\label{eq:corresponding analytic solution}
    B_i^{(t)*} = \left(X_i^T X_i\right)^{-1} X_i^T D_i^{(t)} 
\end{equation}
The resulting local optimal continuous solution is then projected onto the quantization space of a given bit width:
\begin{equation}
    \widetilde{B}_i^{(t)} = \mathcal{Q}\left(B_i^{(t)*}\right) 
\end{equation}
Where $\mathcal{Q}(\cdot)$ is the quantization function. To prevent the block weights from completely jumping to a new quantization level each time without reasonable compensation for the residuals, a step size is introduced $\alpha \in (0,1]$ for the update:
\begin{equation}\label{eq:B}
    B_i^{(t+1)} = B_i^{(t)} + \alpha \left( \widetilde{B}_i^{(t)} - B_i^{(t)} \right)
\end{equation}

This is equivalent to performing a linear interpolation towards the latest solution each time, moving only a portion in each round, similar to the learning rate in gradient descent, thereby allowing the loss to decrease smoothly. This is a process of local minimum quadratic function plus quantization and linear update.

Traditional blockwise greedy quantization strategies suffer from inter-block error accumulation, which causes performance degradation in visually impaired assistance tasks with high stability requirements. To address this problem, this paper designs block based multi-collaborative closed-loop compensation structure based on residual adjustment. In Algorithm \ref{alg:block_residual}, the compensation mechanism is summarized under the guidance of output residual-driven in the RPIQ quantization scheme, which enables subsequent blocks not only to perceive the global error but also to achieve an inter-block collaborative error compensation mechanism based on the residual error of the updated blocks. Through Block based multi-collaborative closed-loop compensation structure based on residual adjustment, RPIQ effectively alleviates the problem of inter-block error accumulation. This improvement in the quantization algorithm will help the output distribution of the quantized model more closely approximate that of the full-precision model, especially in long text reading comprehension and multi-turn question answering tasks, providing a foundation for stable explanations and responses in visual assistance systems.

\begin{algorithm}[t]
    \caption{Block-level residual collaborative compensation}
    \label{alg:block_residual}
    \DontPrintSemicolon 
    
    \SetKwInOut{Input}{Input}
    \SetKwInOut{Output}{Output}
    
    \Input{Full-precision weights $W_{\mathrm{fp}}$, Input information $X$, initial quantized blocks $\{B_i^{\mathrm{init}}\}_{i=1}^M$, Input submatrix $\{X_i\}_{i=1}^M$, Step length $\alpha$.}
    \Output{Refined quantized weight matrix $\{B_i^{\mathrm{init}}\}_{i=1}^M$.}
    \BlankLine
    
    Assemble $W^{\mathrm{init}} \leftarrow [B_1^{\mathrm{init}}, B_2^{\mathrm{init}}, \dots, B_M^{\mathrm{init}}]$\;
    $Y_{\mathrm{orig}} \leftarrow f(X; W_{\mathrm{fp}})$ 
    $Y_q^{\mathrm{init}} \leftarrow f(X; W^{\mathrm{init}})$
    $D \leftarrow Y_{\mathrm{orig}} - Y_q^{\mathrm{init}}$ 
    
    \For{$i = 1, \dots, M$}{
        $Y_q^{\mathrm{init}} \leftarrow f\big(X; [B_1^{\mathrm{init}}, \dots, B_M^{\mathrm{init}}]\big)$\;
        $Y_{q,i}^{(t)} \leftarrow X_i (B_i^{\mathrm{init}})^{\top}$\;
        $D_i^{(t)} \leftarrow Y_{\mathrm{orig}} - \big(Y_q^{(t)} - Y_{q,i}^{(t)}\big)$
        $B_i^{(t)*} \leftarrow (X_i^{\top} X_i)^{-1} X_i^{\top} D_i^{(t)}$ 
        $\tilde{B}_i^{(t)} \leftarrow Q\big(B_i^{(t)*}\big)$ 
        $B_i^{(t+1)} \leftarrow B_i^{\mathrm{init}} + \alpha\big(\tilde{B}_i^{(t)} - B_i^{\mathrm{init}}\big)$ 
    }
\end{algorithm}

\subsection{Single instance calibration based on instantaneous Hessian curvature reconstruction}

\begin{figure*}[htbp]
    \centering
    \includegraphics[width=0.97\textwidth,clip]{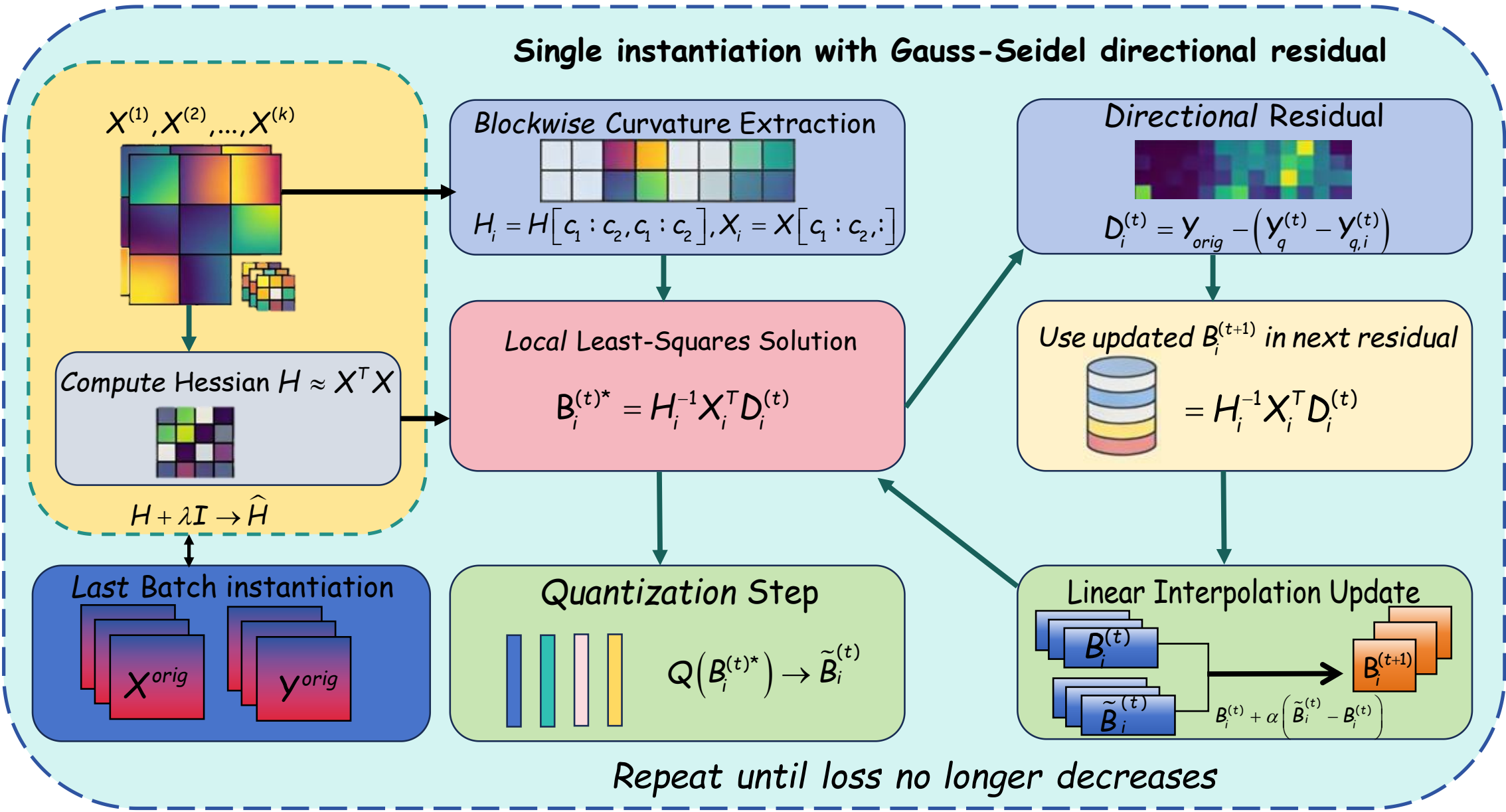}
    \caption{ Single instance calibration paradigm.}
    \label{fig:single_instance_calib}
\end{figure*}

In the practical deployment of visually impaired assistance systems, the quantization process often takes place in resource-constrained environments, making it difficult to retain large-scale calibration data or repeatedly load the global calibration dataset over multiple iterations. To avoid repeatedly traversing the global calibration data while preserving global second-order information, the construction and storage of the global Hessian matrix are completed in the first stage, and in the second stage only the last batch of calibration data is retained for iterative refinement. Specifically, in the first stage of each layer, based on the input features and second-order information accumulated through forward propagation of the calibration data, an approximate Hessian matrix is constructed:
\begin{equation}
    H \approx X^{T} X 
\end{equation}
Where $X$ represents the input feature matrix of this layer collected during the calibration stage, that is, the statistics accumulated across multiple batches of data. Subsequently, in order to enhance the stability of the values, a damping term proportional to the diagonal mean is introduced for $H$:
\begin{equation}
    H + \lambda I \rightarrow \tilde{H}, 
\lambda = \textit{percdamp} \cdot \mathrm{mean}(\mathrm{diag}(H))
\end{equation}
Where $percdamp$ is a hyperparameter used to adjust the damping magnitude, usually a constant, and is used to control the damping amplitude of the $Hessian$ matrix. The function of this hyperparameter is to adjust the magnitude of the damping term to prevent unstable values.

Based on the single instance concept, RPIQ restores the last batch of input and output stored in the calibration stage to:
\begin{equation}
    (X_{\mathrm{orig}}, Y_{\mathrm{orig}}, {H}) = L(W_{\mathrm{fp}}, \textit{percdamp})
\end{equation}
Where $X_{\mathrm{orig}}$ and $Y_{\mathrm{orig}}$ correspond to the last batch of calibration data in the GPTQ calibration process.
$L(\cdot)$ represents the auxiliary function, which is based on the last batch of input and output instances stored during the GPTQ calibration phase.
After entering the second stage, RPIQ no longer accesses the remaining calibration data but always relies on this single batch instance
$(X_ {\mathrm {orig}}, Y_ {\mathrm {orig}}) $, the corresponding output residuals structure; For each block $i$, from a global perspective
The corresponding sub-matrices are extracted from the Hessian matrix:
\begin{equation}
    X_i = X[:, c_1{:}c_2],
H_i = H[c_1{:}c_2, c_1{:}c_2]
\end{equation}

And pre-calculate the approximate inverse curvature corresponding to each block:
\begin{equation}
    H_i^{-1} \approx (X_i^{T} X_i)^{-1}
\end{equation}

Finally, in each round of iteration, a local analytical solution can be obtained:
\begin{equation}
    B_i^{(t)*} = H_i^{-1} X_i^{T} D_i^{(t)}
\end{equation}

In this way, RPIQ does not need to use any other calibration data or rebuild the global Hessian in the second stage
Matrix information, while maintaining the global second-order information, is iteratively optimized only based on this single batch instance. The corresponding single instance calibration process based on the curvature of the $Hessian$ matrix is shown in the Figure \ref{fig:single_instance_calib}.
Rather than concatenating a large amount of calibration data into a huge input tensor during multiple rounds of iteration in memory and video memory
Repeatedly exchange data. If a large amount of calibration data is concatenated, it is very likely to cause OOM. Compared with relying on the entire calibration data:
\begin{equation}
   \mathrm{Memory}_{\text{all}} \approx O\big(\|[X^{(1)}, \ldots, X^{(k)}]\|\big),
\end{equation}
Where $||X\|$ donates the size of a single batch input tensor \(||X^{(1)}, \ldots, X^{(k)}||\) said $k$ batch of calibration data $X ^ {(1)}, \ldots, X ^ {(k)}$ after joining together the tensor of scale.

With memory savings, the memory overhead of the single instance calibration scheme quantitatively proposed by RPIQ is:
\begin{equation}
     \mathrm{Memory}_{\text{RPIQ}} \approx O(\|X\|) \ll \mathrm{Memory}_{\text{all}} 
\end{equation}

Meanwhile, since the quantization scheme relying on fully calibrated data requires multiple data exchanges between memory and video memory, the corresponding time cost will increase significantly. The single instance calibration strategy proposed in this paper will also reduce the time complexity:
\begin{equation}\label{eq:time_complexity}
     \mathrm{Time}_{\text{RPIQ}} \approx O(1)\ll\mathrm{Time}_{all}\approx O(k\cdot T)
\end{equation}
Where \(O(1)\) indicates that the time hardly changes with the batch number and the number of iteration rounds during the iteration process, while $O(k \cdot T)$ indicates that the time increases with the increase of batch number and the number of iteration rounds. The single instance calibration algorithm proposed by RPIQ is as shown Algorithm \ref{alg:single_instance_alog}. 

Given that deploying quantized models in resource-constrained environments for visually impaired assistance systems incurs prohibitive memory and time overhead from repeatedly loading the full calibration dataset across multiple refinement iterations, this paper proposes a single-instance calibration paradigm based on instantaneous Hessian curvature reconstruction. The Single instance calibration paradigm based on Instantaneous Hessian Curvature Reconstruction enables RPIQ to complete multi-round optimization in its second quantization stage by relying only on the last batch of calibration data from the first stage. While retaining global second-order information, this approach controls the extra VRAM (Video RAM) and data loading overhead within an acceptable range. This will be beneficial for deploying large models in resource-limited environments typical of visual assistance systems.

\begin{algorithm}[t]
    \caption{Single instance Hessian-based Calibration}
    \label{alg:single_instance_alog}
    \small 
    \DontPrintSemicolon 
    
    \SetKwInOut{Input}{Input}
    \SetKwInOut{Output}{Output}
    
    \Input{Inputs $\{X^{(j)}\}_{j=1}^{k}$, weights $W_{\mathrm{fp}}$, damping ratio $percdamp$.}
    \Output{$(X_{\mathrm{orig}}, Y_{\mathrm{orig}}, H)$ and block inverses $\{H_i^{-1}\}$.}
    
    \BlankLine
    Initialize $H \leftarrow 0$\;
    
    \For{$b = 1, \dots, k$}{
        $H \leftarrow H + (X^{(b)})^{\top} X^{(b)}$ \tcp*[r]{second-order statistics}
    }
    
    $H \approx X^{\top} X$ \tcp*[r]{Approximate Hessian}
    
    $\lambda \leftarrow \textit{percdamp} \cdot \mathrm{mean}(\mathrm{diag}(H))$ \tcp*[r]{damping coeff.}
    
    $\tilde{H} \leftarrow H + \lambda I$ \tcp*[r]{Apply damping}
    
    \BlankLine
    Set instance to the last calibration batch:\;
     $X_{\mathrm{orig}} \leftarrow X^{(K)}, Y_{\mathrm{orig}} \leftarrow f(X^{(K)}; W_{\mathrm{fp}})$\;
    
    \BlankLine
    \For{each block $i$ with range $[c_1 : c_2]$}{
        Extract blockwise $X_i$ and $H_i$: \;
        \quad $X_i \leftarrow X_{\mathrm{orig}}[:, c_1:c_2], H_i \leftarrow \tilde{H}[c_1:c_2, c_1:c_2]$\;
        $H_i^{-1} \approx (X_i^{\top} X_i)^{-1}$ \tcp*[r]{curvature inverse}
    }
\end{algorithm}

\subsection{Dynamic block iterative quantization based on Gauss-Seidel update}
\begin{figure*}[htbp]
    \centering
    \includegraphics[width=0.97\textwidth,clip]{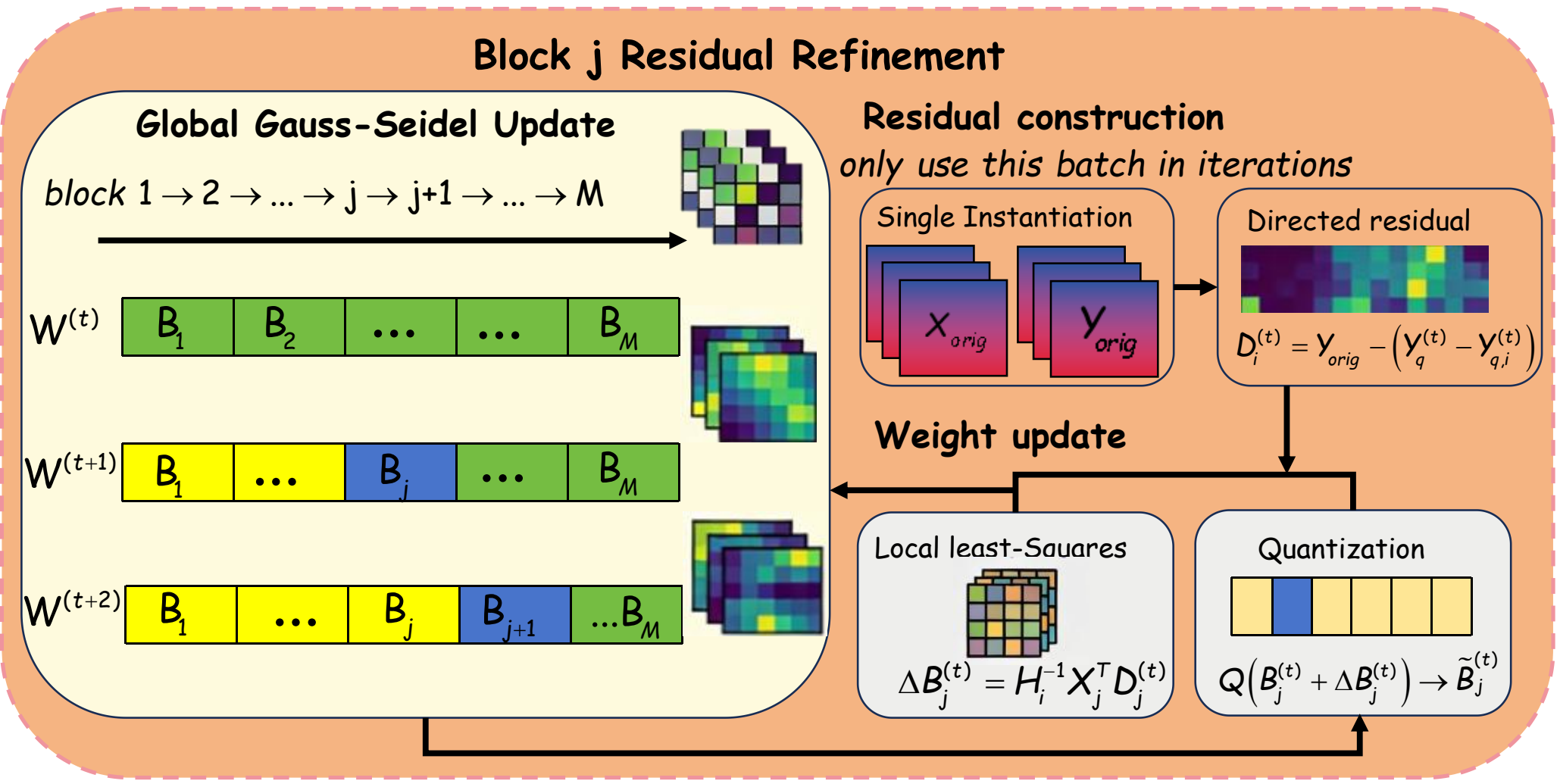}
    \caption{Gauss-Seidel governed dynamic blockwise iterative.}
    \label{fig:Gauss-Seidel}
\end{figure*}

In real-world visually impaired assistance systems requiring high-precision tasks, initial quantization combined with a limited number of compensation iterations may be insufficient to fully reduce all residual errors, potentially leaving precision losses in semantic coherence and emotion understanding. To further improve quantization accuracy and convergence speed while maintaining computational and memory efficiency, RPIQ adopts a Gauss-Seidel style block-level update iteration method in the second stage, dividing the weights of each layer into $M$ column blocks: 
\begin{equation}
    W = [B_1,\ldots,B_M]
\end{equation}

These column blocks together form the weight matrix of this layer. At the $t$-th round of iteration, RPIQ maintains the block weights $\{B_1^{(t)},\ldots,B_M^{(t)}\}$ for this round. When updating the $i$-th block, the network states of its preceding and following blocks in the forward propagation are in a latest-old value mixed state: 
\begin{equation}\label{eq:lastest-old value}
    Y^{(t)} \approx F\! \big(X;\,
B_1^{(t)},\ldots,B_{i-1}^{(t)}, B_i^{(t)},
B_{i+1}^{(t-1)},\ldots,B_M^{(t-1)}\big)
\end{equation}

That is, when updating the $i$th block, the first $i-1$ blocks use the latest weights updated in the current round, while the subsequent blocks use the old values obtained in the previous round. Based on the idea of decomposing the quantization output by blocks, Global output can be obtained corresponding $Y_q^{(t)} = \sum_{j=1}^{M} Y_{q,j}^{(t)}$ and block the output $Y_{q,j}^{(t)} = X_j \big(B_j^{(t)}\big)^{\top}$ for the first among them The input submatrix corresponding to block j, thereby constructing the block-level directional residual:
\begin{equation}
    D_i^{(t)} = Y_{\mathrm{orig}}
- \big( Y_q^{(t)} - Y_{q,i}^{(t)} \big)
\end{equation}

This process is equivalent to subtracting the old contribution rate of the current block from the global residuals, enabling it to better fit the residuals that need to be corrected in this round. Based on this, on the basis of the foregoing account of the least-square solutions $B_ {i} ^ {(t)*}$ and the corresponding quantitative function $Q (\cdot) $, we can get the updated the weight of the current block $B_{i}^{(t+1)}$. After that, when performing calculations, it is only necessary to update the contribution of the current block to the output:
\begin{equation}
    Y_{q,i}^{(t)} = X_i \left( B_{i}^{(t+1)} \right)^T
\end{equation}
\begin{equation}
    Y_{q,i}^{(t+1)} = Y_{q,i}^{(t)} - \left( Y_{q,i}^{(t)} - Y_{q,i}^{(t+1)} \right)
\end{equation}

Based on this, after RPIQ completes the update calculation of the $i$ TH block each time, \(Y_q \) can immediately respond to the latest compensation result of this block. When constructing the residual \(D_{i+1}^{(t)} \) in the next $i+1$ th block, it is based on the network state that already contains all the previously updated blocks. The quantization process is represented by the arrow pointing in the Figure \ref{fig:Gauss-Seidel}, thereby achieving the dynamic block iteration mechanism of Gauss-Seidel update.

At the end of each round, RPIQ will take the output error of a single instance as the optimization objective of the current round and construct the corresponding loss function:
\begin{equation}\label{eq:loss}
    \Gamma^{(t)} = \left\| Y_{orig} - Y_{q}^{(t)} \right\|_2^2
\end{equation}

When the loss function \(\Gamma^{(t)}\) of this round no longer shows any loss decline or the number of iterations \(T_{\max} \) reaches the maximum preset, the machine will be shut down and the quantized weights will be restored to the corresponding optimal solution.

In the practical deployment of visually impaired assistance systems, initial quantization combined with limited compensation iterations alone cannot fully eliminate residual errors, and the model's accuracy in high-performance-demand tasks such as long-text semantic understanding and emotion recognition may still experience significant degradation. To address this, Gauss-Seidel governed dynamic blockwise iterative quantization is introduced, while keeping computational and memory overhead controllable, performs multi-round collaborative refinement on the high-quality initial solution to further improve quantization accuracy and accelerate convergence speed. Since this quantization scheme relies on the GPTQ quantization strategy in the first stage and has already obtained a relatively high-quality initial approximate solution \(W^{\text{init}} \), the Gauss-Seidel update mechanism in the second stage mainly focuses on refining this approximate solution rather than conducting a global search from scratch. The process corresponding to this quantization strategy is shown in Algorithm \ref{alg:gs}. This quantization strategy achieves fast refinement near high-quality approximate solutions. By employing the Gauss-Seidel governed dynamic blockwise iterative update strategy, RPIQ constructs the directional residual for the current block in every iteration based on the latest updated weights, rather than performing optimization in isolation. For intelligent assistive systems for visually impaired users, RPIQ neither increases the inference cost nor compromises the stability of the quantized model's results after deployment, which is beneficial for enhancing the user experience and system reliability.

\begin{algorithm}[t]
    \caption{Dynamic Block Iterative Quantization based on Gauss--Seidel Updates}
    \label{alg:gs}
    \DontPrintSemicolon
    
    \SetKwInOut{Input}{Input}
    \SetKwInOut{Output}{Output}
    
    \Input{Single instance $(X, Y_{\mathrm{orig}})$, initial blocks $\{B_i^{(0)}\}_{i=1}^{M}$, step size $\alpha$, max iterations $T_{\max}$.}
    \Output{Refined quantized blocks $\{B_i^{(t)}\}_{i=1}^{M}$.}
    
    \BlankLine
    Initialize $t \leftarrow 0$\;
    
    \While{$\Gamma^{(t)}$ does not decrease or $t < T_{\max}$}{
        \For{$i = 1$ \KwTo $M$}{
            
            Mixed-state forward: \;
            \quad $Y_q^{(t)} \leftarrow F\big(X; B_1^{(t+1)}, \dots, B_i^{(t)}, \dots, B_M^{(t)}\big)$\;
            
            Block-$i$ contribution: $Y_{q,i}^{(t)} \leftarrow X_i (B_i^{(t)})^{\top}$\;
            
            Directed residual: $D_i^{(t)} \leftarrow Y_{\mathrm{orig}} - (Y_q^{(t)} - Y_{q,i}^{(t)})$\;
            
            Local least-squares: $B_i^{(t)*} \leftarrow (X_i^{\top} X_i)^{-1} X_i^{\top} D_i^{(t)}$\;
            
            Quantization: $\tilde{B}_i^{(t)} \leftarrow Q\big(B_i^{(t)*}\big)$\;
            
            Update: $B_i^{(t+1)} \leftarrow B_i^{(t)} + \alpha\big(\tilde{B}_i^{(t)} - B_i^{(t)}\big)$\;
        }
        Recompute output: $Y_q^{(t+1)} \leftarrow F\big(X; B_1^{(t+1)}, \dots, B_M^{(t+1)}\big)$\;
        
        $\Gamma^{(t+1)} \leftarrow \big\|Y_{\mathrm{orig}} - Y_q^{(t+1)}\big\|_2^2$\;
        
        $t \leftarrow t+1$\;
    }
\end{algorithm}

\section{Experiment}\label{sec:Experiment}
\subsection{Quantization Configuration}
All quantization experiments were conducted on NVIDIA RTX 4090 GPUs with 24GB memory to evaluate the proposed RPIQ method for visually impaired assistance tasks. To ensure reproducibility and fair comparison across all models, fixed calibration datasets were employed for language models and vision-language models respectively.

For language model evaluation, four representative architectures spanning different parameter scales and design philosophies were selected. OPT-6.7B and OPT-13B \cite{zhang2022opt} are decoder-only transformer models trained on diverse web-scale corpora, serving as strong baselines for autoregressive language generation. Qwen3-8B \cite{yang2025qwen3} represents a more recent architecture optimized for multilingual understanding and reasoning capabilities. LLaMA-3.1-8B-Instruct \cite{dubey2024llama} is an instruction-tuned variant optimized for task-oriented dialogue and command execution. The calibration data for language models was sourced from the AllenAI C4 dataset (English subset) \cite{raffel2020exploring, dodge2021documenting}, with 128 samples randomly selected from the streaming dataset and saved as a static file for consistent use across all experiments. All language models were quantized to 4-bit precision using asymmetric quantization with a group size of 128, a widely-adopted standard in recent quantization methods \cite{shao2023omniquant, zhao2024atom, ashkboos2024quarot}, employing the proposed RPIQ method. For the iterative refinement process in RPIQ Stage 2, 5 iterations were performed with an iterative learning rate of 0.01, using only the last sample from the 128 calibration samples to optimize computational efficiency while maintaining quantization quality. All experimental results were compared against the standard GPTQ method \cite{frantar2022gptq} as a baseline to demonstrate the effectiveness of the proposed approach.

For vision-language model evaluation, CogVLM2-19B \cite{hong2024cogvlm2} was selected due to its strong multimodal reasoning capabilities. The calibration data for CogVLM2-19B was sourced from the CogVLM-SFT-311K dataset, with 64 samples used for calibration. The memory overhead during calibration was effectively managed within the 24GB constraint, highlighting the efficiency of the proposed quantization method for practical deployment on consumer-grade hardware. For performance evaluation, RPIQ was integrated as the base quantization method into the cross-modal differentiated quantization framework \cite{wang2025scene}, replacing the original GPTQ-based approach. This framework implements modality-specific strategies to address the varying sensitivity of visual and linguistic components. Similarly, 5 iterations with an iterative learning rate of 0.01 were applied during the iterative refinement process, utilizing the last batch of calibration data instances. Performance comparisons were conducted against the original cross-modal differentiated quantization approach with GPTQ as the base method to validate the improvements achieved by replacing GPTQ with RPIQ in multimodal architectures.

\subsection{Evaluation Methodology}

To comprehensively assess the effectiveness of the proposed RPIQ method, three distinct experimental evaluations were conducted: accuracy assessment, memory overhead analysis, and time cost evaluation.

The accuracy evaluation was performed across multiple benchmarks to measure both language understanding and multimodal reasoning capabilities. For language models, two complementary metrics were employed: perplexity and downstream task accuracy. In the context of assistive technologies for visually impaired users, language models serve as the cognitive backbone for processing and generating natural language descriptions of visual content, answering user queries about their surroundings, and facilitating human-computer interaction through conversational interfaces. Perplexity measures the model's uncertainty in predicting the next token, serving as a fundamental indicator of language modeling capability that directly impacts the fluency and coherence of generated descriptions. Following standard practice \cite{xiao2023smoothquant, tseng2024quip, dettmers2023spqr}, the evaluation was conducted on the WikiText-2 dataset \cite{merity2016pointer}, specifically using the raw version (wikitext-2-raw-v1) test split containing 4,358 sequences. 

The perplexity computation follows the implementation provided by the AutoGPTQ library \cite{autogptq}. For each evaluation batch, the model's cross-entropy loss is computed and averaged over the tokens within that batch. The final perplexity is then calculated by averaging these batch-level losses across all evaluation batches and taking the exponential:
\begin{equation}
\text{PPL} = \exp\left(\frac{1}{N}\sum_{i=1}^{N}\mathcal{L}_i\right)
\end{equation}
where $N$ is the number of batches and $\mathcal{L}_i$ represents the average cross-entropy loss for batch $i$. Lower perplexity values indicate better language modeling performance, ensuring that the quantized models maintain sufficient linguistic capabilities for generating coherent scene descriptions and responding to user inquiries in assistive scenarios.

To evaluate performance on practical downstream tasks relevant to visually impaired assistance, a sentiment classification benchmark was employed using tweet data \cite{rosenthal2017semeval}. Sentiment understanding represents a crucial capability for assistive systems that need to interpret the emotional tone of textual information encountered in the user's environment, such as social media posts, product reviews, or public announcements, thereby providing contextually appropriate assistance. The task requires categorizing input text into three sentiment classes: negative, neutral, and positive. A total of 870 test samples were evaluated using the prompt template ``Question: What's the sentiment of the given text? Choices are \{labels\}. Text: \{text\} Answer:'', where $\{\text{labels}\} = \{\text{negative}, \text{neutral}, \text{positive}\}$ and $\{\text{text}\}$ denotes the input tweet content. The classification accuracy is calculated as:
\begin{equation}
\text{Accuracy} = \frac{1}{N}\sum_{i=1}^{N}\mathbb{1}[\hat{y}_i = y_i]
\end{equation}
where $N=870$ is the total number of test samples, $\hat{y}_i$ is the predicted label, $y_i$ is the ground truth label, and $\mathbb{1}[\cdot]$ is the indicator function. This metric directly reflects the model's capability in real-world text understanding scenarios that are essential for comprehensive assistive functionalities.

For vision-language models, the OCR-VQA benchmark \cite{mishra2019ocr} was adopted to assess multimodal reasoning capabilities requiring both visual perception and text comprehension. This benchmark is particularly relevant to visually impaired assistance applications, as the ability to recognize and comprehend text embedded within images—such as street signs, product labels, document contents, and storefront information—constitutes a fundamental assistive service that enables independent navigation and daily activity management. OCR-VQA comprises 207,572 book cover images with over 1 million question-answer pairs, demanding the model to read text within images and answer natural language questions accordingly. The evaluation was conducted on the OCR-VQA-TESTCORE subset, which provides a representative sample of the full benchmark for efficient evaluation. Given an image $\mathbf{I}$ and a question $\mathbf{q}$, the model generates an answer $\hat{a}$ which is compared against the ground truth answer $a$ using exact match accuracy:
\begin{equation}
\text{Accuracy}_{\text{OCR-VQA}} = \frac{1}{|\mathcal{T}|}\sum_{(\mathbf{I}, \mathbf{q}, a) \in \mathcal{T}}\mathbb{1}[\hat{a}(\mathbf{I}, \mathbf{q}) = a]
\end{equation}
where $\mathcal{T}$ denotes the test set and $\hat{a}(\mathbf{I}, \mathbf{q})$ represents the model's predicted answer. This benchmark particularly challenges quantized models due to the requirement of preserving fine-grained text recognition capabilities within compressed parameters, which is critical for ensuring that quantization does not compromise the practical utility of assistive systems in real-world scenarios. For all accuracy evaluations, baseline comparisons were established against GPTQ \cite{frantar2022gptq} for language-only models, and against the cross-modal differentiated quantization framework with GPTQ as the base method \cite{wang2025scene} for vision-language models. Additionally, full-precision model performance was measured to characterize the performance variation introduced by quantization, as certain architectural properties may lead to either degradation or improvement in specific metrics under compression.

To validate the practical feasibility of the proposed single instance refinement approach, peak memory consumption during the quantization process was monitored and recorded. The memory overhead $\Delta M$ introduced by the iterative refinement procedure was quantified by comparing the peak GPU memory usage between RPIQ and the baseline GPTQ method:
\begin{equation}
\Delta M = M_{\text{RPIQ}}^{\text{peak}} - M_{\text{GPTQ}}^{\text{peak}}
\end{equation}
where $M_{\text{RPIQ}}^{\text{peak}}$ and $M_{\text{GPTQ}}^{\text{peak}}$ represent the maximum memory consumption observed during the quantization process using RPIQ and GPTQ respectively. Through the single instance refinement strategy that processes only the last calibration sample during iterative optimization, the additional memory requirements imposed by maintaining gradient information and performing iterative updates were effectively controlled, demonstrating that the proposed method introduces acceptable overhead relative to the baseline approach for consumer-grade hardware.

The temporal efficiency of the proposed method was evaluated by measuring the total quantization time required to process the entire model. The absolute time difference $\Delta T$ between RPIQ and GPTQ was computed as:
\begin{equation}
\Delta T = T_{\text{RPIQ}} - T_{\text{GPTQ}}
\end{equation}
where $T_{\text{RPIQ}}$ and $T_{\text{GPTQ}}$ represent the total quantization time using RPIQ and GPTQ respectively. This metric directly quantifies the additional computational cost in absolute terms introduced by the iterative refinement mechanism, demonstrating that the time increase remains modest across different model scales and validating that the convergence occurs efficiently within the allocated 5 iterations.

\subsection{Quantization Accuracy Evaluation}

\subsubsection{Language Model Performance}

The quantization accuracy was evaluated on four representative language models across perplexity and sentiment classification benchmarks. Table \ref{tab:language_model_results} presents the comprehensive performance comparison between full-precision models (BF16), GPTQ-quantized models (4-bit), and RPIQ-quantized models (4-bit).

\begin{table*}[htbp]
\centering
\caption{Performance Comparison of Language Models Under Different Quantization Methods}
\label{tab:language_model_results}
\resizebox{\textwidth}{!}{
\begin{threeparttable}
\begin{tabular}{@{}lccc|ccc|ccc@{}}
\toprule[2pt]
\multirow{2}{*}{Model} & \multicolumn{3}{c|}{BF16} & \multicolumn{3}{c|}{GPTQ (4-bit)} & \multicolumn{3}{c}{RPIQ (4-bit, Ours)} \\
\cmidrule(lr){2-4} \cmidrule(lr){5-7} \cmidrule(lr){8-10}
& Acc (\%) & PPL & Mem (GB) & Acc (\%) & PPL & Mem (GB) & Acc (\%) & PPL & Mem (GB) \\
\midrule
OPT-6.7B & 44.25 & 24.058 & 13.4 & 39.66 & 24.397 & 3.91 & \textbf{40.34} & \textbf{24.393} & 3.91 \\
OPT-13B & OOM\tnote{*} & OOM\tnote{*} & 26 & 32.76 & 24.646 & 6.69 & \textbf{35.52} & \textbf{24.579} & 6.69 \\
Qwen3-8B & 55.40 & 17.890 & 16 & 55.52 & 19.166 & 5.75 & \textbf{56.09} & \textbf{18.889} & 5.75 \\
LLaMA-3.1-8B-Instruct & 63.22 & 16.942 & 16 & 59.89 & 17.661 & 5.94 & \textbf{63.56} & 17.885 & 5.94 \\
\bottomrule[2pt]
\end{tabular}
\begin{tablenotes}
\small
\item[*] Out of memory on 24GB GPU for full-precision inference
\end{tablenotes}
\end{threeparttable}
}
\end{table*}

The experimental results demonstrate the effectiveness of RPIQ in preserving language model capabilities under 4-bit quantization, which is particularly critical for visually impaired assistance applications where linguistic coherence and semantic understanding directly impact user experience. The evaluation spans four architecturally diverse models: OPT-6.7B and OPT-13B, representing the decoder-only transformer family trained on diverse web-scale corpora~\cite{zhang2022opt}; Qwen3-8B, a state-of-the-art multilingual foundation model with enhanced reasoning capabilities and cross-lingual transfer learning~\cite{yang2025qwen3}; and LLaMA-3.1-8B-Instruct, an instruction-tuned variant specifically optimized for following user commands and task-oriented dialogue~\cite{dubey2024llama}. This diverse model selection ensures comprehensive validation across different architectural paradigms, training objectives, and deployment scenarios relevant to assistive technology applications.

The memory footprint analysis reveals substantial compression benefits for both quantization methods. Compared to full-precision models requiring 13.4--38 GB of GPU memory, the 4-bit quantization reduces memory consumption to 3.91--11.3 GB, enabling deployment of sophisticated language models on consumer-grade hardware with limited computational resources. This compression efficiency is particularly significant for the OPT-13B model, which becomes practically deployable after quantization despite exceeding available memory at full precision.

Across all evaluated architectures, RPIQ consistently demonstrates superior performance compared to the baseline GPTQ method on sentiment classification accuracy while achieving competitive or improved perplexity scores. For the OPT family models, RPIQ achieves enhanced accuracy on both OPT-6.7B and OPT-13B compared to GPTQ. These accuracy gains are accompanied by simultaneous perplexity improvements, demonstrating that RPIQ's iterative refinement mechanism successfully enhances both task-specific performance and general language modeling quality concurrently. These improvements in sentiment understanding capability are essential for assistive systems that need to accurately interpret the emotional context of environmental text, enabling appropriate responses when reading social media content, product reviews, or public announcements to users.

The perplexity analysis reveals a particularly compelling advantage of RPIQ's residual-projected iterative refinement approach. For three out of four evaluated models (OPT-6.7B, OPT-13B, and Qwen3-8B), RPIQ achieves lower perplexity than GPTQ, with Qwen3-8B demonstrating substantial perplexity reduction. This consistent perplexity improvement across multiple architectures validates that RPIQ's block based multi-collaborative compensation mechanism effectively mitigates the quantization-induced information loss that accumulates in GPTQ's greedy one-shot optimization. The single instance calibration paradigm based on Hessian curvature reconstruction enables RPIQ to preserve the global second-order statistical information while performing targeted iterative corrections, thereby achieving perplexity values that more closely approximate or even surpass the original full-precision models' linguistic coherence. The relative perplexity variation from full-precision models remains tightly controlled for models where full-precision evaluation was feasible, ensuring that the fluency of generated scene descriptions and conversational responses remains largely indistinguishable from unquantized models in practical deployment scenarios.

The performance on Qwen3-8B demonstrates particularly compelling results for practical deployment in multilingual assistive scenarios. RPIQ achieves enhanced sentiment accuracy, surpassing both GPTQ and the full-precision baseline, while simultaneously recovering perplexity closer to the full-precision reference. This simultaneous enhancement in both metrics suggests that the proposed residual propagation and Gauss-Seidel iterative refinement mechanisms effectively mitigate quantization-induced information loss for this architecture. Given Qwen3-8B's advanced multilingual capabilities and cross-lingual transfer learning design, the preserved semantic understanding is crucial for assistive applications requiring accurate interpretation of user queries and environmental text across diverse linguistic contexts, from reading foreign-language signage to processing multilingual documents. The ability to match or exceed full-precision performance in downstream task accuracy while achieving competitive perplexity indicates that RPIQ preserves not only the model's linguistic fluency but also its deeper semantic reasoning capabilities across languages.

For LLaMA-3.1-8B-Instruct, RPIQ achieves remarkable accuracy preservation compared to GPTQ, effectively recovering to the full-precision level. As an instruction-tuned model specifically optimized for task-oriented dialogue and command execution, this substantial gain demonstrates that RPIQ's optimization strategy is particularly effective at preserving the specialized capabilities encoded during instruction fine-tuning. The instruction-tuning process aligns the model's representations with human intent and task-oriented reasoning patterns, making these capabilities especially valuable yet potentially vulnerable to quantization-induced degradation. RPIQ's block based multi-collaborative compensation and iterative refinement mechanisms successfully maintain these fine-tuned alignments, enabling the quantized model to retain its ability to comprehend user commands, interpret situational context, and generate appropriate task-oriented responses. For visually impaired users relying on voice-activated assistive interfaces that require accurate interpretation of spoken commands and situational queries, this preservation of instruction-tuned capabilities translates directly to enhanced system responsiveness and task completion reliability in real-world interactive scenarios.

The consistent superiority of RPIQ across diverse architectural designs—including decoder-only transformers (OPT), multilingual foundation models with advanced reasoning (Qwen3), and instruction-tuned variants (LLaMA-3.1-Instruct)---validates the generalizability of the proposed method for practical assistive technology deployment. The accuracy improvements on sentiment classification directly translate to enhanced capability in interpreting contextual information from the user's environment, while the maintained or improved perplexity scores ensure that generated descriptions and conversational responses remain fluent and coherent. For visually impaired users relying on these systems for environmental understanding and daily task assistance, these quantitative improvements represent tangible enhancements in system reliability and communicative quality. Furthermore, the successful quantization of OPT-13B, which exceeds memory capacity at full precision on consumer-grade hardware, demonstrates that RPIQ enables deployment of larger, more capable models within resource constraints typical of personal assistive devices, thereby expanding the accessibility of advanced language understanding capabilities to users who would otherwise be limited to smaller models.

\subsubsection{Vision-Language Model Performance}

The quantization effectiveness was further evaluated on CogVLM2-19B for multimodal reasoning tasks requiring integrated visual perception and language comprehension. To assess the effectiveness of the proposed RPIQ method in multimodal architectures, the GPTQ base method within the cross-modal differentiated quantization framework (CMDQ)~\cite{wang2025scene} was replaced with the proposed RPIQ method. Given the larger parameter scale and multimodal architecture complexity, 64 calibration samples were used to ensure the quantization process remained feasible within the 24GB GPU memory constraint typical of consumer-grade hardware, demonstrating the practical applicability of the proposed approach. Table~\ref{tab:vlm_results} presents the OCR-VQA benchmark performance across different book categories, comparing the original full-precision model, the CMDQ baseline (using GPTQ as the base method), and the CMDQ+RPIQ variant (using RPIQ as the base method) with varying iteration counts. Note that the table shows representative category-specific results, while the overall accuracy reflects performance across the complete evaluation set.

\begin{table*}[htbp]
\centering
\caption{OCR-VQA Performance Comparison on CogVLM2-19B Under Different Quantization Configurations}
\label{tab:vlm_results}
\resizebox{\textwidth}{!}{
\begin{threeparttable}
\begin{tabular}{@{}lcc|ccccc@{}}
\toprule[2pt]
Method & Overall & Mem (GB) & Cookbooks & Medical & History & Reference & Education \\ 
\midrule
CogVLM2-19B-Chat (Original)\tnote{*} & 64.90 & 38 & 68.80 & 61.50 & 65.60 & 54.20 & 59.40 \\
\midrule
CMDQ (4-bit) & 63.00 & 11.3 & 65.63 & 55.20 & 68.75 & 51.04 & 56.25 \\
CMDQ + RPIQ (4-bit, 5 iter) & \textbf{63.70} & 11.3 & \textbf{67.70} & \textbf{59.38} & 68.75 & \textbf{53.13} & \textbf{62.50} \\
CMDQ + RPIQ (4-bit, 20 iter) & 58.17 & 11.3 & 61.46 & 55.21 & 68.75 & 38.54 & 55.21 \\
\bottomrule[2pt]
\end{tabular}
\begin{tablenotes}
\small
\item[*] All performance data is sourced from the official VLMEvalKit~\cite{duan2024vlmevalkit} leaderboard.
\item Note: All accuracy values are reported as percentages. CMDQ refers to the cross-modal differentiated quantization framework with GPTQ as the base method. The table presents representative category results; overall accuracy is computed across the complete benchmark.
\end{tablenotes}
\end{threeparttable}
}
\end{table*}

The memory footprint analysis demonstrates substantial compression benefits achieved through 4-bit quantization. The original CogVLM2-19B-Chat model requires 38 GB of GPU memory for full-precision inference, which is reduced to 11.3 GB when applying both CMDQ and RPIQ methods. This compression ratio enables deployment of sophisticated vision-language models on consumer-grade hardware, making advanced multimodal assistive capabilities accessible on devices with limited computational resources.

Performance comparison with the original model reveals that applying 4-bit quantization introduces moderate accuracy degradation across most categories, reflecting the inherent trade-off between model compression and task performance. Category-specific analysis shows varied sensitivity to quantization: the History category exhibits enhanced performance in quantized variants compared to the original model, suggesting potential regularization effects, while domains such as Medical and Reference demonstrate more pronounced accuracy reduction. These category-specific variations reflect the differential impact of quantization on diverse visual-linguistic reasoning patterns encountered across different text domains.

The experimental results reveal that integrating RPIQ into the cross-modal differentiated quantization framework yields measurable improvements in multimodal reasoning capabilities, which are indispensable for assistive applications requiring visual text recognition and comprehension. When applying 5 iterations of refinement, the RPIQ-enhanced approach achieves enhanced overall accuracy on OCR-VQA compared to the CMDQ baseline. More importantly, the performance gains are particularly pronounced in domain-specific categories critical for practical assistance scenarios. The Cookbooks category experiences notable improvement, the Medical category shows substantial gain, and the Education category demonstrates significant enhancement. These category-specific improvements directly impact real-world assistive functionalities: enhanced cookbook recognition enables better assistance with cooking instructions and ingredient identification, improved medical text comprehension supports reading prescription labels and health information, and strengthened educational material processing facilitates access to learning resources and documentation.

The consistent performance on the History category across quantization methods suggests that certain domain characteristics may exhibit inherent robustness to quantization variations, potentially due to the stylistic consistency and vocabulary patterns typical of historical texts. However, the Reference category shows modest improvement with RPIQ, indicating that while gains are achieved across most domains, the magnitude of improvement varies with the semantic complexity and visual diversity present in different text categories. For visually impaired users, these improvements translate to more reliable text recognition when interacting with diverse printed materials encountered in daily life—from following recipe instructions to understanding medication information to accessing educational content—thereby enhancing independence in activities that fundamentally depend on accurate visual text comprehension.

The performance comparison between 5-iteration and 20-iteration refinement configurations reveals a critical insight regarding the single instance refinement strategy employed in RPIQ. While the 5-iteration configuration demonstrates consistent improvements across most categories, extending the iterative process to 20 iterations results in substantial performance degradation, with overall accuracy declining considerably and particularly severe drops observed in the Reference category. This degradation pattern strongly suggests the occurrence of overfitting during the extended refinement process. When iterative optimization is conducted using a single calibration instance as implemented in the proposed approach for memory efficiency—the model's quantization parameters become excessively specialized to the specific characteristics of that individual sample. The limited diversity inherent in single instance refinement fails to provide sufficient regularization against overfitting, causing the quantization parameters to capture idiosyncratic patterns specific to the refinement instance rather than generalizable statistical properties of the broader data distribution. This phenomenon is particularly detrimental in multimodal contexts where visual-linguistic relationships exhibit high variance across different image-text combinations.

The sharp decline in Reference and Education categories under 20 iterations indicates that these domains may be especially susceptible to overfitting, possibly due to their heterogeneous content structure and diverse visual layouts compared to more standardized formats in History texts. From a practical deployment perspective, these results validate the design choice of limiting refinement to 5 iterations as implemented in the proposed method. The 5-iteration configuration successfully balances the benefits of iterative error correction with the risk of single instance overfitting, achieving meaningful accuracy improvements while maintaining generalization across diverse content domains. For assistive technology applications targeting visually impaired users, this finding underscores the importance of calibrating the refinement process to prevent model degradation that would compromise reliability in real-world usage scenarios where content diversity far exceeds calibration data. The ability to preserve text recognition accuracy across cookbooks, medical documents, and educational materials with the properly tuned 5-iteration approach ensures that quantized assistive systems maintain the breadth of capability necessary for comprehensive daily assistance, from reading recipes and ingredient labels to understanding medication instructions to accessing learning materials.

\subsubsection{Qualitative Analysis}

\begin{figure*}[htbp]
	\centering
	\includegraphics[width=0.95\textwidth]{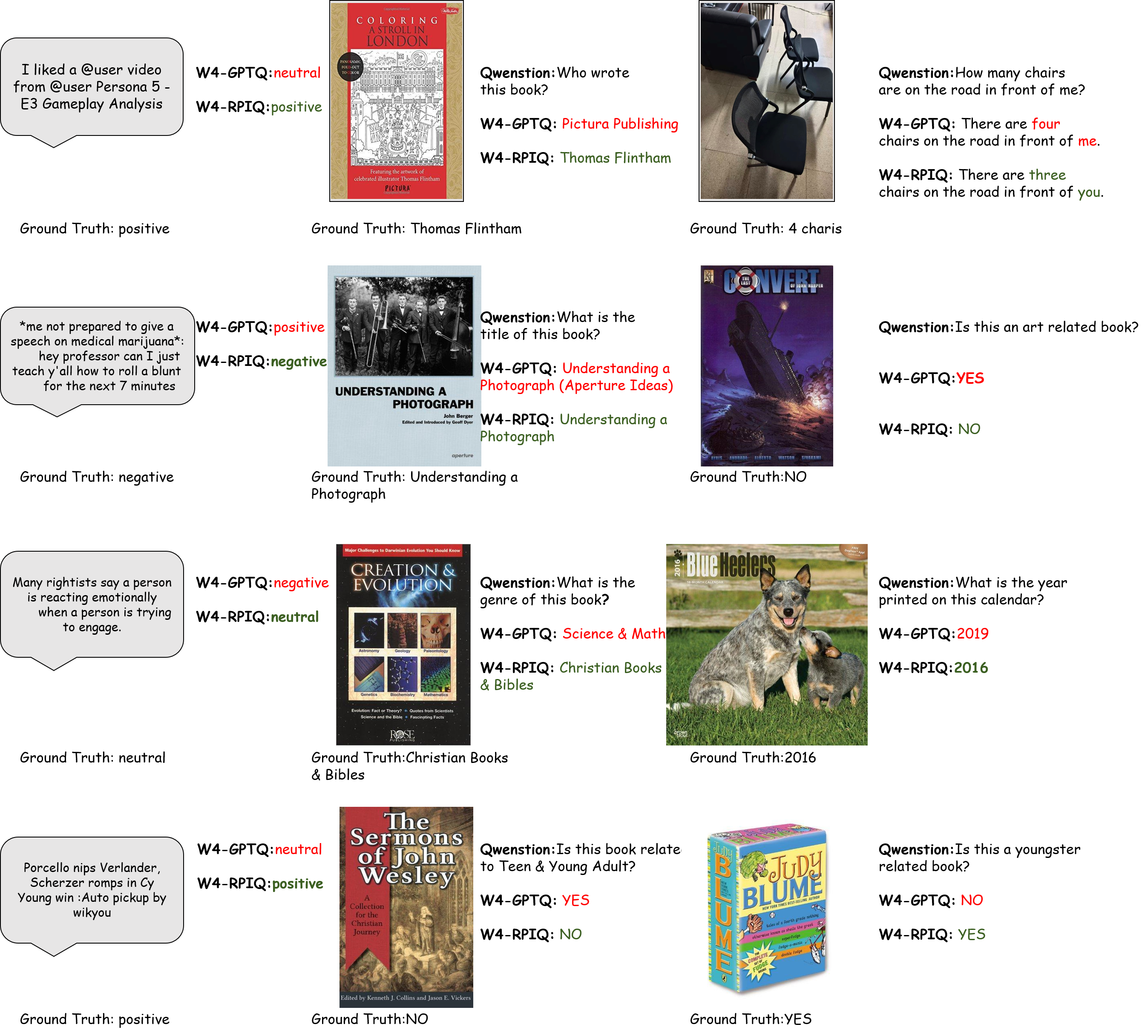}
	\caption{Qualitative results of quantized models on representative tasks. The proposed method significantly improves the semantic understanding and visual perception accuracy compared to the GPTQ baseline. Correct predictions are colored in green and wrong predictions in red.}
	\label{fig:qualitative_showcases}
\end{figure*}

Beyond quantitative metrics, qualitative examples are presented demonstrating RPIQ's superior preservation of semantic accuracy across diverse task modalities. Figure \ref{fig:qualitative_showcases} illustrates multiple representative cases comparing GPTQ and RPIQ on sentiment classification, OCR-based visual question answering, and visual assistance tasks.

For sentiment analysis tasks, RPIQ demonstrates more precise emotion recognition capabilities, correctly distinguishing between positive, neutral, and negative sentiments, while GPTQ exhibits sentiment polarity misclassifications across multiple cases—a distinction critical for assistive systems interpreting user emotional context. In OCR-VQA tasks, RPIQ shows superior text recognition and information extraction abilities, accurately identifying book authors, titles, genres, and other key information, whereas GPTQ frequently produces information confusion or misidentification, directly impacting the reliability of reading assistance systems. In visual assistance tasks, RPIQ reveals significant advantages in spatial reasoning, such as accurately counting the number of chairs in object counting tasks while GPTQ produces erroneous counts; moreover, RPIQ demonstrates higher accuracy in year recognition and book classification tasks.

These qualitative improvements indicate that RPIQ not only outperforms baseline methods in statistical metrics but also provides more reliable semantic understanding and visual perception capabilities in practical application scenarios. Such enhancements are particularly important for assistive technology contexts, where the accuracy of model outputs directly determines the effectiveness and experiential quality for visually impaired users navigating complex multimodal information. Whether interpreting emotional tones in social media text, extracting accurate information from book covers, or identifying the number of objects in surrounding environments, RPIQ consistently demonstrates higher accuracy and robustness.

\subsection{Memory Overhead Analysis}

To validate the practical feasibility of the proposed single instance calibration paradigm during the quantization process, the peak GPU memory usage was monitored and recorded for both the baseline GPTQ method and the RPIQ approach across all evaluated models. Table~\ref{tab:memory_overhead} presents the comprehensive memory consumption comparison, quantifying the additional overhead introduced by the iterative refinement mechanism while maintaining single instance calibration efficiency.

\begin{table}[htbp]
\centering
\begin{threeparttable}
\caption{Peak Memory Consumption During Quantization Process}
\label{tab:memory_overhead}
\begin{tabular}{lccc}
\toprule
\textbf{Model} & \textbf{GPTQ} & \textbf{RPIQ} & \textbf{$\Delta M$} \\
 & \textbf{(GB)} & \textbf{(GB)} & \textbf{(GB)} \\
\midrule
OPT-6.7B & 10.04 & 11.08 & +1.04 \\
OPT-13B & 14.62 & 16.50 & +1.88 \\
Qwen3-8B & 5.94 & 7.32 & +1.38 \\
LLaMA-3.1-8B-Instruct & 6.80 & 9.41 & +2.61 \\
CogVLM2-19B & 15.36 & 20.63 & +5.27 \\
\bottomrule
\end{tabular}
\begin{tablenotes}
\small
\item Note: CogVLM2-19B was calibrated using 64 samples, while language models used 128 samples.
\end{tablenotes}
\end{threeparttable}
\end{table}

The experimental results demonstrate that the additional memory overhead introduced by RPIQ remains within acceptable bounds across all evaluated architectures, validating the effectiveness of the single instance refinement strategy. For language models, the absolute memory increase ranges from 1.04 GB (OPT-6.7B) to 2.61 GB (LLaMA-3.1-8B-Instruct), representing relative overhead ratios between 10.4\% and 38.4\% compared to the baseline GPTQ method. The vision-language model CogVLM2-19B, calibrated with 64 samples, exhibits the largest absolute increase of 5.27 GB, corresponding to a 34.3\% relative overhead. This elevated overhead reflects both the inherent complexity of multimodal architectures, which necessitate joint optimization of visual and linguistic components during iterative refinement, and the substantially larger parameter count at the 19B-parameter scale.

Despite the iterative nature of the proposed refinement mechanism, which performs 5 rounds of block based multi-collaborative compensation, the memory overhead remains substantially lower than what would be incurred by methods requiring full calibration dataset retention. The single instance calibration paradigm based on Hessian curvature reconstruction enables RPIQ to maintain only the last batch of calibration data ($X_{\text{orig}}$, $Y_{\text{orig}}$) and the pre-computed global Hessian matrix $H$ in memory during Stage 2 iterations, rather than concatenating multiple calibration batches into large tensor structures. This design choice effectively prevents the Out-of-Memory (OOM) errors that frequently plague quantization methods requiring $\mathcal{O}(\|[X^{(1)}, \ldots, X^{(k)}]\|)$ memory allocation for processing $k$ calibration batches simultaneously. Furthermore, by maintaining a single instance approach, RPIQ avoids the repeated data loading and exchange between system memory and GPU memory that would be necessary for methods requiring multiple passes over the calibration dataset. While such repeated data loading primarily incurs time overhead rather than memory overhead, the architectural decision to process only the last calibration batch contributes to both memory efficiency and computational efficiency. As will be demonstrated in the subsequent time cost evaluation, this design achieves the accuracy improvements documented in preceding sections while introducing minimal additional temporal overhead.

The memory overhead analysis reveals consistent scaling behavior across different model sizes and architectures. The absolute memory increase grows approximately proportionally with model parameter count, indicating that the iterative refinement mechanism does not introduce pathological memory scaling characteristics. This predictable scaling property enables reliable estimation of hardware requirements for quantizing models of varying sizes. The ability to quantize OPT-13B (16.50 GB peak) and CogVLM2-19B (20.63 GB peak) within the 24 GB constraint of consumer-grade GPUs demonstrates that RPIQ maintains practical applicability across model scales relevant to visually impaired assistance applications.

For assistive systems targeting visually impaired users, the demonstrated memory efficiency has practical implications for deployment scenarios. The controlled memory footprint ensures that quantization can be performed on widely available consumer-grade hardware without requiring specialized high-memory infrastructure. For applications deployed in privacy-sensitive contexts, such as reading personal documents or interpreting private communications, the ability to perform quantization locally eliminates the need to transmit sensitive calibration data to external servers, thereby enhancing user privacy while maintaining the accuracy improvements validated in preceding sections. The combination of acceptable memory overhead and preserved model accuracy establishes RPIQ as a practical quantization solution for real-world assistive technology deployment scenarios serving visually impaired users.

\subsection{Time Cost Evaluation}\label{se:Time cost}

To assess the computational efficiency of the proposed iterative refinement mechanism, the total quantization time was measured across all evaluated models. Table~\ref{tab:time_cost} presents the comprehensive temporal overhead comparison between GPTQ and RPIQ, quantifying the additional time introduced by the multi-round block based multi-collaborative compensation strategy while maintaining practical deployment feasibility.

\begin{table}[htbp]
\centering
\caption{Total Quantization Time Comparison}
\label{tab:time_cost}
\begin{tabular}{lccc}
\hline
Model & GPTQ (s) & RPIQ (s) & $\Delta T$ (s) \\
\hline
OPT-6.7B & 376.94 & 389.35 & +12.41 \\
OPT-13B & 665.47 & 683.92 & +18.45 \\
Qwen3-8B & 431.54 & 450.47 & +18.93 \\
LLaMA-3.1-8B-Instruct & 417.72 & 488.40 & +70.68 \\
CogVLM2-19B & 1609.89 & 1907.06 & +297.17 \\
\hline
\end{tabular}
\end{table}

The experimental results demonstrate that the additional time overhead introduced by RPIQ's iterative refinement mechanism remains within acceptable bounds for practical deployment scenarios. For language models ranging from 6.7B to 13B parameters, the absolute time increase varies from 12.41 seconds (OPT-6.7B) to 18.93 seconds (Qwen3-8B), corresponding to less than one-third of a GPU minute of additional processing time. Even for the larger LLaMA-3.1-8B-Instruct model, which exhibits the highest language model overhead at 70.68 seconds, the additional temporal cost amounts to approximately 1.2 GPU minutes for offline quantization processes.

The vision-language model CogVLM2-19B demonstrates the largest absolute time increase at 297.17 seconds (approximately 5 GPU minutes), reflecting both the substantially larger parameter count at the 19B scale and the inherent complexity of multimodal architectures requiring joint optimization of visual and linguistic components during iterative refinement.

The controlled time overhead across all evaluated architectures validates the efficiency of the single instance calibration paradigm based on Hessian curvature reconstruction. By maintaining only the last batch of calibration data $(X_{orig}, Y_{orig})$ and the pre-computed global Hessian matrix $H$ during Stage 2 iterations, RPIQ avoids the substantial temporal overhead that would be incurred by methods requiring repeated data loading and forward propagation through multiple calibration batches. The temporal efficiency is mathematically formalized in Equation~(\ref{eq:time_complexity}), where $\text{Time}_{\text{RPIQ}} \approx O(1)$ indicates that the additional time overhead remains approximately constant with respect to the number of calibration batches, in contrast to methods requiring $O(k \cdot T)$ complexity for processing $k$ batches across $T$ iterations.

Furthermore, the observed time overhead exhibits predictable scaling characteristics across different model sizes. The absolute time increase grows approximately proportionally with model parameter count and architectural complexity. The OPT series demonstrates this pattern clearly, with OPT-13B requiring 18.45 seconds compared to OPT-6.7B's 12.41 seconds, maintaining a consistent overhead ratio relative to model scale.

For visually impaired assistance applications, the minimal additional processing time measured in seconds for 8B-parameter models and minutes for 19B-parameter models validates the computational efficiency of the proposed approach on consumer-grade hardware.

\subsection{Convergence Analysis}

To validate the effectiveness of the iterative refinement mechanism and demonstrate the convergence behavior of the proposed RPIQ method, the evolution of the loss function $\Gamma^{(t)} = \|Y_{\text{orig}} - Y_q^{(t)}\|_2^2$ defined in Equation \ref{eq:loss} was systematically monitored across multiple iterations during the quantization process. This loss function quantifies the Frobenius norm of output residuals between the full-precision output and the quantized output at iteration $t$, serving as the optimization objective for the block based multi-collaborative compensation strategy. The convergence analysis was conducted across representative layers from both language models and vision-language models, providing comprehensive insight into the optimization dynamics.

\begin{figure*}[htbp]
\centering
\begin{subfigure}[b]{0.48\textwidth}
    \centering
    \includegraphics[width=\textwidth]{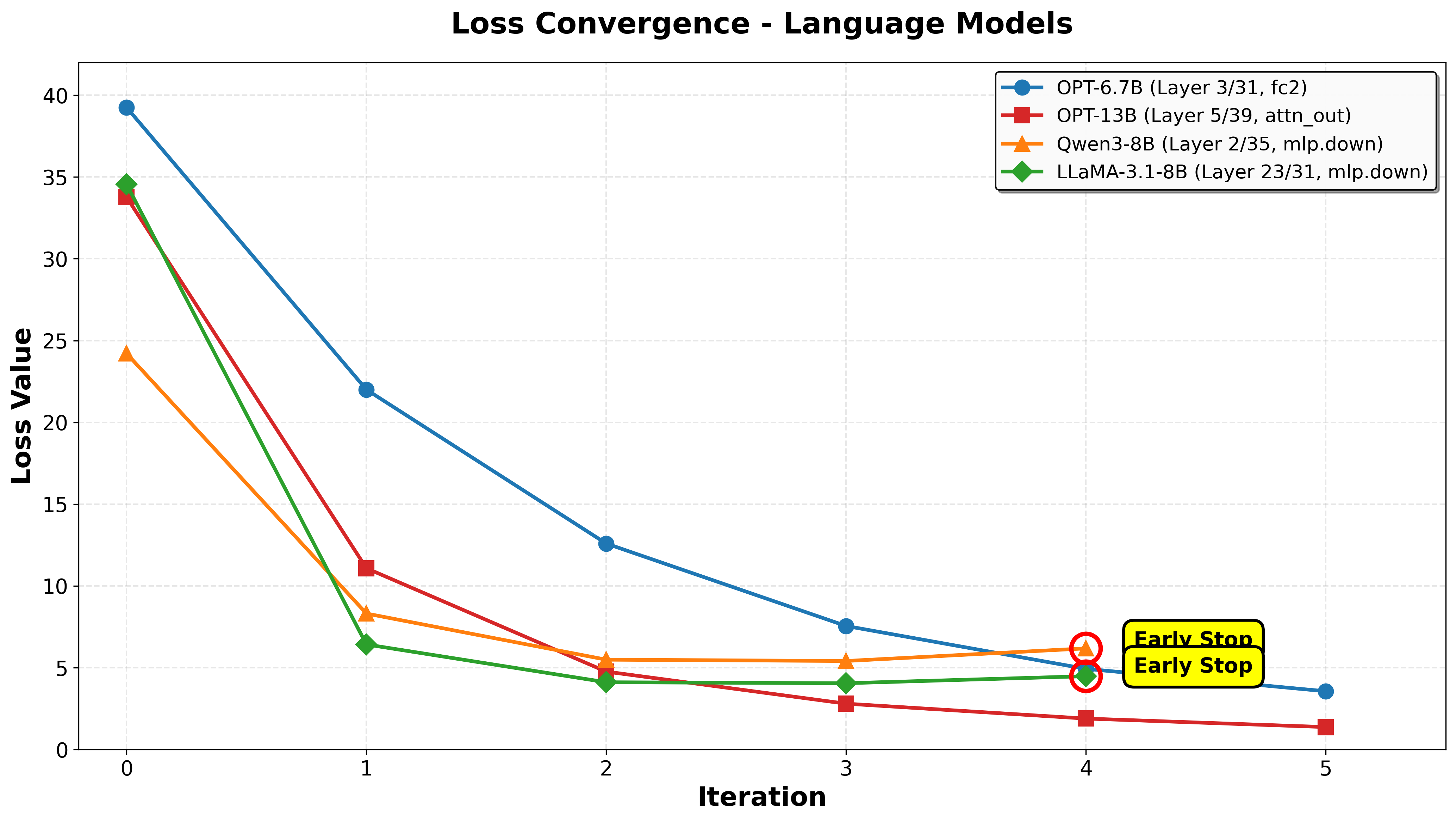}
    \caption{Language model comparison across architectures}
    \label{fig:convergence_language}
\end{subfigure}
\hfill
\begin{subfigure}[b]{0.48\textwidth}
    \centering
    \includegraphics[width=\textwidth]{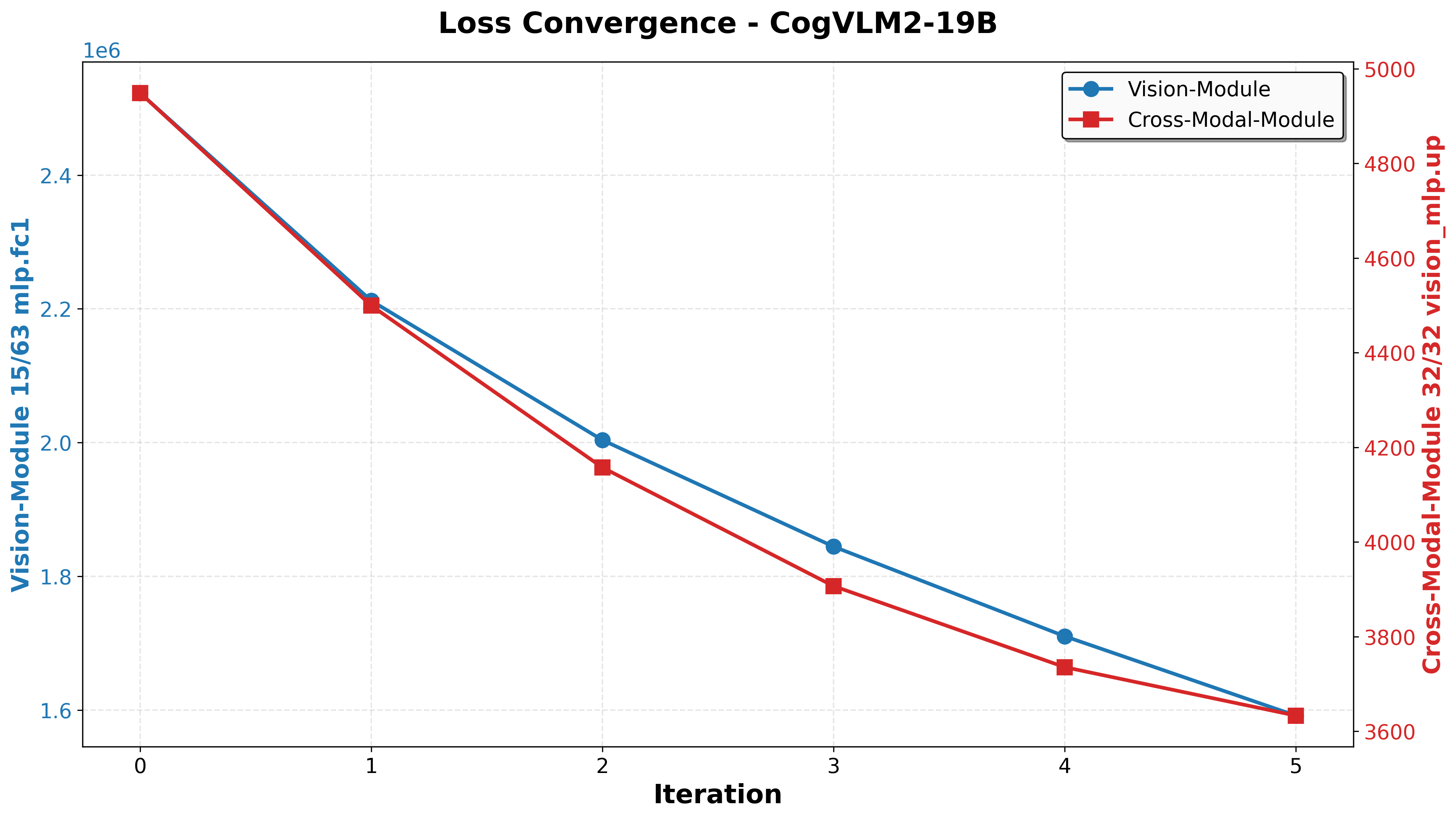}
    \caption{CogVLM2-19B vision and cross-modal modules}
    \label{fig:convergence_cogvlm}
\end{subfigure}
\caption{Loss convergence trajectories across representative model components during RPIQ Stage 2 iterative refinement, where iteration 0 represents the initial loss after GPTQ quantization (Stage 1). (a) Language models exhibit diverse convergence patterns, with Qwen3-8B and LLaMA-3.1-8B-Instruct achieving early stopping at iteration 4 when the convergence criteria in Algorithm \ref{alg:gs} are satisfied. (b) CogVLM2-19B demonstrates rapid convergence in both vision and cross-modal modules within 5 iterations.}
\label{fig:convergence}
\end{figure*}

Figure \ref{fig:convergence} presents the loss convergence trajectories for representative model components, where iteration 0 represents the initial loss $\Gamma^{(0)}$ after GPTQ quantization in Stage 1, before RPIQ refinement begins. Figure \ref{fig:convergence}(a) compares the convergence patterns across four language models (OPT-6.7B, OPT-13B, Qwen3-8B, and LLaMA-3.1-8B-Instruct), revealing diverse convergence characteristics across different architectures. Notably, both Qwen3-8B and LLaMA-3.1-8B-Instruct achieve convergence at iteration 4 as indicated by the "Early Stop" markers, demonstrating that the loss function $\Gamma^{(t)}$ satisfied the convergence criteria defined in Algorithm \ref{alg:gs} without exhausting the maximum 5-iteration budget. Figure \ref{fig:convergence}(b) illustrates the convergence behavior of CogVLM2-19B's vision and cross-modal modules, where both components achieve substantial loss reduction within 5 iterations. The Vision Module exhibits rapid initial descent from iteration 0 to iteration 5, achieving 36.90\% total reduction, while the Cross-Modal Module demonstrates steady convergence representing 26.58\% improvement.

The quantitative convergence statistics are summarized in Table \ref{tab:convergence}, presenting detailed loss reduction metrics for representative layers across all evaluated models. The Initial Loss column corresponds to $\Gamma^{(0)}$ immediately after Stage 1 GPTQ quantization, while Final Loss represents $\Gamma^{(t_{\text{final}})}$ at convergence or maximum iterations. The results demonstrate consistent convergence behavior with total loss reductions ranging from 26.58\% to 95.93\% across different model components. Language models exhibit particularly strong convergence performance, with OPT-13B achieving 95.93\% loss reduction in the self-attention output projection layer (layer 5 of 39) and OPT-6.7B attaining 90.92\% reduction in the fully connected layer (layer 3 of 31). According to Algorithm \ref{alg:gs}, the iterative process terminates when $\Gamma^{(t)}$ ceases to decrease or when the maximum iteration count is reached. The observed early stopping at iteration 4 for Qwen3-8B (77.66\% reduction) and LLaMA-3.1-8B-Instruct (88.27\% reduction in layer 23 of 31) indicates that the loss function satisfied the convergence criteria before completing all 5 iterations, thereby avoiding unnecessary computational overhead while achieving substantial output residual minimization.

For the vision-language model CogVLM2-19B, the convergence characteristics differ substantially between modality-specific components due to their distinct computational properties and parameter scales. The Vision Module's fully connected layer achieves 36.90\% loss reduction over 5 iterations, with absolute loss values orders of magnitude larger than language components, reflecting the different scales of visual versus linguistic feature representations. The Cross-Modal Module's vision MLP up-projection layer demonstrates 26.58\% reduction over the full 5 iterations, indicating that the optimization process continued to reduce $\Gamma^{(t)}$ throughout the entire iteration budget without triggering early stopping.

The early stopping behavior observed in Qwen3-8B and LLaMA-3.1-8B-Instruct provides empirical validation of the convergence criteria implemented in Algorithm \ref{alg:gs} (line 2). When the loss function ceases to demonstrate meaningful reduction between consecutive iterations, the optimization process terminates automatically at iteration 4, avoiding unnecessary computational overhead while achieving near-optimal quantization quality. This adaptive termination mechanism ensures computational efficiency across diverse architectures without requiring manual hyperparameter tuning for each model family.

The convergence patterns reveal several important characteristics of the proposed RPIQ method's optimization dynamics. First, the majority of loss reduction occurs within the initial 2-4 iterations for most architectures, with diminishing returns in subsequent iterations. This rapid initial convergence is attributable to the block based multi-collaborative compensation mechanism defined in Equations \ref{eq:d}-\ref{eq:B}, where the directed residual $D_i^{(t)} = Y_{\text{orig}} - (Y_q^{(t)} - Y_{q,i}^{(t)})$ constructed for each block in Equation \ref{eq:d} provides a precise correction target. Starting from GPTQ's approximate solution in Stage 1 ensures that the initial residual $\Gamma^{(0)}$ is already substantially reduced compared to random initialization, enabling the local least-squares optimization in Equation \ref{eq:target} to efficiently refine the quantization within a favorable parameter region. Second, the controlled iteration count (4-5 iterations) maintains practical computational efficiency while achieving substantial reductions in the global loss, as evidenced by the time cost analysis in Section \ref{se:Time cost}. The Gauss-Seidel update strategy expressed in Equation \ref{eq:lastest-old value} enables each block $i$ to immediately utilize the refined weights $\{B_1^{(t+1)}, \ldots, B_{i-1}^{(t+1)}\}$ from previously optimized blocks within the same iteration, thereby propagating corrections forward through the block sequence and accelerating convergence compared to isolated block optimization.

The convergence analysis demonstrates that the block based multi-collaborative compensation mechanism successfully reduces the output residual loss across diverse model architectures and components. The Gauss-Seidel iterative update strategy formalized in Algorithm \ref{alg:gs} enables efficient convergence by constructing the mixed-state forward propagation in Equation \ref{eq:lastest-old value}, where each block $i$ leverages the latest refined weights from blocks $1$ to $i-1$ when solving its local least-squares problem in Equation \ref{eq:corresponding analytic solution}. This contrasts with GPTQ's one-shot blockwise optimization, where each block is quantized in isolation without the opportunity to compensate for errors accumulated in previous blocks. The directed residual construction in Equation \ref{eq:d} ensures that each block's optimization target $D_i^{(t)}$ accurately reflects the global output error while accounting for the current contributions of all other blocks, thereby enabling coordinated error reduction across the entire weight matrix. For visually impaired assistance applications, this reliable convergence behavior ensures that quantized models achieve stable and predictable performance improvements without requiring extensive hyperparameter tuning or architecture-specific modifications, facilitating practical deployment across diverse assistive technology platforms.

\begin{table*}[htbp]
\centering
\caption{Convergence Statistics for Representative Layers Across Evaluated Models}
\label{tab:convergence}
\resizebox{0.95\textwidth}{!}{
\begin{threeparttable}
\begin{tabular}{@{}llcccccc@{}}
\toprule[2pt]
Model & Component & Layer & Initial Loss & Final Loss & Total Reduction & Reduction (\%) & Iterations \\ 
\midrule
OPT-6.7B & fc2 & 3 of 31 & 39.25 & 3.56 & 35.68 & 90.92 & 5 \\
OPT-13B & attn\_out & 5 of 39 & 33.77 & 1.37 & 32.40 & 95.93 & 5 \\
Qwen3-8B & mlp.down & 2 of 35 & 24.21 & 5.41 & 18.80 & 77.66 & 4$^{\dagger}$ \\
LLaMA-3.1-8B-Instruct & mlp.down & 23 of 31 & 34.55 & 4.05 & 30.50 & 88.27 & 4$^{\dagger}$ \\
\midrule
CogVLM2-Vision & mlp.fc1 & Vision Module & 2,522,746.50 & 1,591,786.25 & 930,960.25 & 36.90 & 5 \\
CogVLM2-Cross & mlp.vision\_mlp.up & Cross-Modal Module & 4,948.58 & 3,633.26 & 1,315.32 & 26.58 & 5 \\
\bottomrule[2pt]
\end{tabular}
\begin{tablenotes}
\small
\item Note: Loss values represent the output residual loss $\Gamma^{(t)} = \|Y_{\text{orig}} - Y_q^{(t)}\|_2^2$ defined in Equation \ref{eq:loss}. Iteration 0 represents the initial loss $\Gamma^{(0)}$ after GPTQ quantization (Stage 1) before RPIQ refinement begins. The Layer column indicates the specific layer within the total layer count for each architecture. $^{\dagger}$Early stopping was triggered at iteration 4 when the convergence criteria in Algorithm \ref{alg:gs} were satisfied.
\end{tablenotes}
\end{threeparttable}
}
\end{table*}

\section{Discussion}\label{sec:Discussion}
This chapter presents a systematic discussion of the research results and the RPIQ quantization strategy developed in this work. First, it provides a detailed exposition of the experimental data to demonstrate the advances achieved in model optimization and application performance. It then analyzes the key methodological components that enable these improvements and examines the inherent limitations of the current single-instance calibration paradigm. Finally, the chapter distills the main insights of this study and discusses their implications for the efficient deployment and practical implementation of large-scale models.

\subsection{Research Results}
The experimental results of this study demonstrate that the proposed RPIQ quantization strategy successfully addresses the trade-off between accuracy and efficiency in Post-Training Quantization (PTQ). Regarding model performance, this method significantly reduces the inter-block error accumulation caused by the traditional greedy quantization in GPTQ. Specifically, on the LLaMA-3.1-8B-Instruct model, RPIQ achieves a 3.67\%   accuracy improvement in sentiment classification tasks compared to standard GPTQ, effectively enhancing quantization performance. In the multimodal domain, the CogVLM2-19B model quantized via RPIQ yields an overall improvement of 0.70\% on the OCR-VQA benchmark, with a substantial gain of as much as 6.25\% observed in the Education category.

Crucially, these accuracy enhancements are achieved while maintaining high computational efficiency. By adopting the single-instance calibration paradigm, the increase in peak memory overhead is constrained within the range of 10.4\% to 38.4\%, and the additional time cost for the 19B model is limited to approximately 5 minutes. This strongly confirms the feasibility of deploying high-precision iterative refinement schemes on consumer-grade hardware, such as the RTX 4090, without the need for industrial-scale computing clusters.

\subsection{Methodological Advantages}
At the methodological level, RPIQ provides a set of mutually coordinated improvements that specifically target three key issues of GPTQ. First, to address the problem of unidirectional inter-block error accumulation and convergence to local optima, RPIQ builds global output residuals and blockwise directional residuals on top of the GPTQ initial solution, and performs multi-round, block based multi-cooperative closed-loop compensation via local least-squares solutions, quantization projection, and linear interpolation updates, so that blocks to be quantized can, during their updates, perceive and exploit the errors corrected by previously updated blocks, thereby mitigating the one-way accumulation of errors across blocks.

Second, to address the repeated calibration of global data and the risk of OOM, RPIQ uses all calibration data in the first stage to construct an approximate Hessian matrix, while in the second stage it retains only the input-output instance of the last batch together with the corresponding blockwise curvature information, and performs multi-round refinement via a single-instance Hessian curvature reconstruction paradigm, thereby preserving global second-order statistics while reducing the data requirement during the iterative phase to a single-batch scale. Furthermore, RPIQ adopts a Gauss-Seidel style dynamic blockwise iteration strategy, in which each iteration round consistently uses a mixed network state consisting of already-quantized and not-yet-quantized blocks for forward propagation and directional residual construction, ensuring that the quantization of each block is based on the latest global residual and enabling the method to obtain a better quantization solution with fewer iteration rounds.

\subsection{Limitations and Methological Considerations}
Despite the significant advantages demonstrated by the proposed quantization framework, a comprehensive understanding of its performance requires an examination of its limitations and the implications of methodological choices.

The proposed quantization framework by this paper carries an inherent risk of single-instance overfitting, and a critical limitation observed in the ablation experiments is the sensitivity of the model to the number of iterations. When the refinement process is extended to 20 iterations, a marked decline in model performance occurs, as evidenced by a 12.50\% drop in the Reference category.

This phenomenon suggests that excessive iterations relying on a single instance may lead to overfitting, whereby model parameters over-fit the noise patterns of that specific instance rather than the general distribution of the data.

Furthermore, the proposed quantization framework exhibits a significant dependency on initialization, as the RPIQ method, serving as a refinement process, relies on the high-quality initial solutions provided by GPTQ. Although the designed scheme achieves rapid convergence, typically within 4 to 5 iterations, this characteristic implies that the final results are inevitably constrained by the attraction domain of the initial values. If the initial quantization point is too distant from the optimum, the local least-squares updates may fail to locate the global optimal solution.

\subsection{Insights and Implications}
The research results show that, through block based multi-collaborative closed-loop compensation, single instance calibration, and Gauss-Seidel governed dynamic blockwise iteration, RPIQ offers an innovative path that extends GPTQ from one-shot greedy quantization to multi-round collaborative refinement under controllable resource constraints. This path demonstrates that, without substantially increasing resource consumption in the quantization phase, it is possible in low-bit quantization of large models to simultaneously pursue higher accuracy, better robustness, and lower deployment cost, thereby providing valuable insights for applying large model compression algorithms on assistive devices for visually impaired users. 

For intelligent assistive systems designed for visually impaired users, the improvements brought by RPIQ on tasks such as language understanding, sentiment analysis, and OCR‑VQA can be directly leveraged to enhance the interpretation and feedback mechanisms for diverse information sources, including social media text, medication instructions, and educational materials, thereby strengthening the applicability of assistive systems in real-world scenarios.

\section{Conclusion}\label{sec:Conclusion}
This study successfully addresses the high-precision challenges of deploying large-scale models in resource-constrained assistive scenarios. Through the block-based multi-collaborative closed-loop compensation structure under residual adjustment and the Gauss-Seidel governed dynamic blockwise iterative quantization strategy, significant accuracy improvements are achieved, restoring the performance of LLaMA-3.1 to near-full-precision levels and enhancing the fine-grained text recognition capabilities of CogVLM2-19B. Most importantly, the designed single-instance calibration paradigm demonstrates that high-quality iterative optimization does not necessitate the repeated loading of full datasets. Quantization of a 19B-parameter model is successfully completed on a single consumer-grade GPU with acceptable time and memory overhead, making the deployment of advanced visual impairment assistance systems on personal devices technically feasible.

Based on these foundations, future work will primarily unfold in several key directions. An automated dynamic snapshot selection mechanism is intended to be developed to periodically rotate calibration data in memory without increasing peak memory overhead, thereby providing better regularization effects. Simultaneously, the integration of the RPIQ quantization strategy with cross-modal differentiated quantization strategies will be explored to further expand the boundaries of efficiency and accuracy. Finally, there are plans to extend the RPIQ quantization strategy to multimodal models encompassing audio, video, and 3D point clouds. Combined with low-level instruction set optimizations for specific mobile NPUs or embedded computing platforms, these efforts will facilitate the implementation of second-level quantization refinement in practical applications such as assisted driving and real-time interactive devices for the disabled.

\bibliographystyle{unsrt}
\bibliography{reference}

@article{dettmers20218,
  title={8-bit optimizers via block-wise quantization},
  author={Dettmers, Tim and Lewis, Mike and Shleifer, Sam and Zettlemoyer, Luke},
  journal={arXiv preprint arXiv:2110.02861},
  year={2021}
}

@inproceedings{guo2024gptqt,
  title={GPTQT: quantize large language models twice to push the efficiency},
  author={Guo, Yipin and Lang, Yilin and Ren, Qinyuan},
  booktitle={2024 IEEE International Conference on Cybernetics and Intelligent Systems (CIS) and IEEE International Conference on Robotics, Automation and Mechatronics (RAM)},
  pages={368--373},
  year={2024},
  organization={IEEE}
}

@article{frantar2022gptq,
  title={Gptq: Accurate post-training quantization for generative pre-trained transformers},
  author={Frantar, Elias and Ashkboos, Saleh and Hoefler, Torsten and Alistarh, Dan},
  journal={arXiv preprint arXiv:2210.17323},
  year={2022}
}

@article{nahshan2021loss,
  title={Loss aware post-training quantization},
  author={Nahshan, Yury and Chmiel, Brian and Baskin, Chaim and Zheltonozhskii, Evgenii and Banner, Ron and Bronstein, Alex M and Mendelson, Avi},
  journal={Machine Learning},
  volume={110},
  number={11},
  pages={3245--3262},
  year={2021},
  publisher={Springer}
}

@article{lin2024awq,
  title={Awq: Activation-aware weight quantization for on-device llm compression and acceleration},
  author={Lin, Ji and Tang, Jiaming and Tang, Haotian and Yang, Shang and Chen, Wei-Ming and Wang, Wei-Chen and Xiao, Guangxuan and Dang, Xingyu and Gan, Chuang and Han, Song},
  journal={Proceedings of machine learning and systems},
  volume={6},
  pages={87--100},
  year={2024}
}

@incollection{gholami2022survey,
  title={A survey of quantization methods for efficient neural network inference},
  author={Gholami, Amir and Kim, Sehoon and Dong, Zhen and Yao, Zhewei and Mahoney, Michael W and Keutzer, Kurt},
  booktitle={Low-power computer vision},
  pages={291--326},
  year={2022},
  publisher={Chapman and Hall/CRC}
}

@article{esser2019learned,
  title={Learned step size quantization},
  author={Esser, Steven K and McKinstry, Jeffrey L and Bablani, Deepika and Appuswamy, Rathinakumar and Modha, Dharmendra S},
  journal={arXiv preprint arXiv:1902.08153},
  year={2019}
}

@inproceedings{nagel2020up,
  title={Up or down? adaptive rounding for post-training quantization},
  author={Nagel, Markus and Amjad, Rana Ali and Van Baalen, Mart and Louizos, Christos and Blankevoort, Tijmen},
  booktitle={International conference on machine learning},
  pages={7197--7206},
  year={2020},
  organization={PMLR}
}

@article{li2021brecq,
  title={Brecq: Pushing the limit of post-training quantization by block reconstruction},
  author={Li, Yuhang and Gong, Ruihao and Tan, Xu and Yang, Yang and Hu, Peng and Zhang, Qi and Yu, Fengwei and Wang, Wei and Gu, Shi},
  journal={arXiv preprint arXiv:2102.05426},
  year={2021}
}

@article{yao2022zeroquant,
  title={Zeroquant: Efficient and affordable post-training quantization for large-scale transformers},
  author={Yao, Zhewei and Yazdani Aminabadi, Reza and Zhang, Minjia and Wu, Xiaoxia and Li, Conglong and He, Yuxiong},
  journal={Advances in neural information processing systems},
  volume={35},
  pages={27168--27183},
  year={2022}
}

@inproceedings{nagel2019data,
  title={Data-free quantization through weight equalization and bias correction},
  author={Nagel, Markus and Baalen, Mart van and Blankevoort, Tijmen and Welling, Max},
  booktitle={Proceedings of the IEEE/CVF international conference on computer vision},
  pages={1325--1334},
  year={2019}
}

@inproceedings{wang2019haq,
  title={Haq: Hardware-aware automated quantization with mixed precision},
  author={Wang, Kuan and Liu, Zhijian and Lin, Yujun and Lin, Ji and Han, Song},
  booktitle={Proceedings of the IEEE/CVF conference on computer vision and pattern recognition},
  pages={8612--8620},
  year={2019}
}

@article{hubara2020improving,
  title={Improving post training neural quantization: Layer-wise calibration and integer programming},
  author={Hubara, Itay and Nahshan, Yury and Hanani, Yair and Banner, Ron and Soudry, Daniel},
  journal={arXiv preprint arXiv:2006.10518},
  year={2020}
}

@article{zhang2025comq,
  title={Comq: A backpropagation-free algorithm for post-training quantization},
  author={Zhang, Aozhong and Yang, Zi and Wang, Naigang and Qi, Yingyong and Xin, Jack and Li, Xin and Yin, Penghang},
  journal={IEEE Access},
  year={2025},
  publisher={IEEE}
}

@article{li2025gptaq,
  title={GPTAQ: Efficient Finetuning-Free Quantization for Asymmetric Calibration},
  author={Li, Yuhang and Yin, Ruokai and Lee, Donghyun and Xiao, Shiting and Panda, Priyadarshini},
  journal={arXiv preprint arXiv:2504.02692},
  year={2025}
}

@article{yao2022rapq,
  title={Rapq: Rescuing accuracy for power-of-two low-bit post-training quantization},
  author={Yao, Hongyi and Li, Pu and Cao, Jian and Liu, Xiangcheng and Xie, Chenying and Wang, Bingzhang},
  journal={arXiv preprint arXiv:2204.12322},
  year={2022}
}

@inproceedings{hubara2021accurate,
  title={Accurate post training quantization with small calibration sets},
  author={Hubara, Itay and Nahshan, Yury and Hanani, Yair and Banner, Ron and Soudry, Daniel},
  booktitle={International conference on machine learning},
  pages={4466--4475},
  year={2021},
  organization={PMLR}
}

@article{nagel2106white,
  title={A white paper on neural network quantization. arXiv 2021},
  author={Nagel, Markus and Fournarakis, Marios and Amjad, Rana Ali and Bondarenko, Yelysei and van Baalen, Mart and Blankevoort, Tijmen},
  journal={arXiv preprint arXiv:2106.08295},
  volume={4}
}

@article{hadi2023large,
  title={Large language models: a comprehensive survey of its applications, challenges, limitations, and future prospects},
  author={Hadi, Muhammad Usman and Qureshi, Rizwan and Shah, Abbas and Irfan, Muhammad and Zafar, Anas and Shaikh, Muhammad Bilal and Akhtar, Naveed and Wu, Jia and Mirjalili, Seyedali and others},
  journal={Authorea preprints},
  volume={1},
  number={3},
  pages={1--26},
  year={2023},
  publisher={Authorea}
}

@article{li2024efficient,
  title={Efficient LLMs training and inference: An introduction},
  author={Li, Rui and Fu, Deji and Shi, Chunyu and Huang, Zhilan and Lu, Gang},
  journal={IEEE Access},
  year={2024},
  publisher={IEEE}
}

@inproceedings{holiel2024assisting,
  title={Assisting visually impaired subjects using large language models: a comprehensive evaluation},
  author={Holiel, Heidi Ahmed and Fawzi, Sahar Ali and Al-Atabany, Walid},
  booktitle={2024 6th Novel Intelligent and Leading Emerging Sciences Conference (NILES)},
  pages={561--566},
  year={2024},
  organization={IEEE}
}

@inproceedings{leporini2025preliminary,
  title={A Preliminary Evaluation of Generative AI Tools for Blind Users: Usability and Screen Reader Interaction},
  author={Leporini, Barbara and Buzzi, Marina and Della Penna, Giuseppe},
  booktitle={Proceedings of the 18th ACM International Conference on PErvasive Technologies Related to Assistive Environments},
  pages={562--568},
  year={2025}
}

@article{shao2023omniquant,
  title={Omniquant: Omnidirectionally calibrated quantization for large language models},
  author={Shao, Wenqi and Chen, Mengzhao and Zhang, Zhaoyang and Xu, Peng and Zhao, Lirui and Li, Zhiqian and Zhang, Kaipeng and Gao, Peng and Qiao, Yu and Luo, Ping},
  journal={arXiv preprint arXiv:2308.13137},
  year={2023}
}

@article{souza2024intelligent,
  title={Intelligent environments and assistive technologies for assisting visually impaired people: a systematic literature review},
  author={Souza, Leandro Rossetti de and Francisco, Rosemary and Rosa Tavares, Jo{\~a}o Elison da and Barbosa, Jorge Luis Vict{\'o}ria},
  journal={Universal Access in the Information Society},
  pages={1--28},
  year={2024},
  publisher={Springer}
}

@article{abidi2024comprehensive,
  title={A comprehensive review of navigation systems for visually impaired individuals},
  author={Abidi, Mustufa Haider and Siddiquee, Arshad Noor and Alkhalefah, Hisham and Srivastava, Vishwaraj},
  journal={Heliyon},
  volume={10},
  number={11},
  year={2024},
  publisher={Elsevier}
}

@article{zhao2024vialm,
  title={Vialm: A survey and benchmark of visually impaired assistance with large models},
  author={Zhao, Yi and Zhang, Yilin and Xiang, Rong and Li, Jing and Li, Hillming},
  journal={arXiv preprint arXiv:2402.01735},
  year={2024}
}

@inproceedings{li2025vocot,
  title={Vocot: Unleashing visually grounded multi-step reasoning in large multi-modal models},
  author={Li, Zejun and Luo, Ruipu and Zhang, Jiwen and Qiu, Minghui and Huang, Xuan-Jing and Wei, Zhongyu},
  booktitle={Proceedings of the 2025 Conference of the Nations of the Americas Chapter of the Association for Computational Linguistics: Human Language Technologies (Volume 1: Long Papers)},
  pages={3769--3798},
  year={2025}
}

@article{zhou2024lidar,
  title={Lidar-ptq: Post-training quantization for point cloud 3d object detection},
  author={Zhou, Sifan and Li, Liang and Zhang, Xinyu and Zhang, Bo and Bai, Shipeng and Sun, Miao and Zhao, Ziyu and Lu, Xiaobo and Chu, Xiangxiang},
  journal={arXiv preprint arXiv:2401.15865},
  year={2024}
}

@inproceedings{xie2025beyond,
  title={Beyond Visual Perception: Insights from Smartphone Interaction of Visually Impaired Users with Large Multimodal Models},
  author={Xie, Jingyi and Yu, Rui and Zhang, He and Billah, Syed Masum and Lee, Sooyeon and Carroll, John M},
  booktitle={Proceedings of the 2025 CHI Conference on Human Factors in Computing Systems},
  pages={1--17},
  year={2025}
}

@article{baig2024ai,
  title={AI-based wearable vision assistance system for the visually impaired: Integrating real-time object recognition and contextual understanding using large vision-language models},
  author={Baig, Mirza Samad Ahmed and Gillani, Syeda Anshrah and Shah, Shahid Munir and Aljawarneh, Mahmoud and Khan, Abdul Akbar and Siddiqui, Muhammad Hamzah},
  journal={arXiv preprint arXiv:2412.20059},
  year={2024}
}

@article{wang2025scene,
  title={Scene-Aware Vectorized Memory Multi-Agent Framework with Cross-Modal Differentiated Quantization VLMs for Visually Impaired Assistance},
  author={Wang, Xiangxiang and Wang, Xuanyu and Luo, YiJia and Yu, Yongbin and Fan, Manping and Zhang, Jingtao and Ren, Liyong},
  journal={Expert Systems with Applications},
  pages={130662},
  year={2025},
  publisher={Elsevier}
}

@article{raffel2020exploring,
  title={Exploring the limits of transfer learning with a unified text-to-text transformer},
  author={Raffel, Colin and Shazeer, Noam and Roberts, Adam and Lee, Katherine and Narang, Sharan and Matena, Michael and Zhou, Yanqi and Li, Wei and Liu, Peter J},
  journal={Journal of machine learning research},
  volume={21},
  number={140},
  pages={1--67},
  year={2020}
}

@article{dodge2021documenting,
  title={Documenting large webtext corpora: A case study on the colossal clean crawled corpus},
  author={Dodge, Jesse and Sap, Maarten and Marasovi{\'c}, Ana and Agnew, William and Ilharco, Gabriel and Groeneveld, Dirk and Mitchell, Margaret and Gardner, Matt},
  journal={arXiv preprint arXiv:2104.08758},
  year={2021}
}

@article{zhao2024atom,
  title={Atom: Low-bit quantization for efficient and accurate llm serving},
  author={Zhao, Yilong and Lin, Chien-Yu and Zhu, Kan and Ye, Zihao and Chen, Lequn and Zheng, Size and Ceze, Luis and Krishnamurthy, Arvind and Chen, Tianqi and Kasikci, Baris},
  journal={Proceedings of Machine Learning and Systems},
  volume={6},
  pages={196--209},
  year={2024}
}

@article{ashkboos2024quarot,
  title={Quarot: Outlier-free 4-bit inference in rotated llms},
  author={Ashkboos, Saleh and Mohtashami, Amirkeivan and Croci, Maximilian L and Li, Bo and Cameron, Pashmina and Jaggi, Martin and Alistarh, Dan and Hoefler, Torsten and Hensman, James},
  journal={Advances in Neural Information Processing Systems},
  volume={37},
  pages={100213--100240},
  year={2024}
}

@article{hong2024cogvlm2,
  title={Cogvlm2: Visual language models for image and video understanding},
  author={Hong, Wenyi and Wang, Weihan and Ding, Ming and Yu, Wenmeng and Lv, Qingsong and Wang, Yan and Cheng, Yean and Huang, Shiyu and Ji, Junhui and Xue, Zhao and others},
  journal={arXiv preprint arXiv:2408.16500},
  year={2024}
}

@inproceedings{xiao2023smoothquant,
  title={Smoothquant: Accurate and efficient post-training quantization for large language models},
  author={Xiao, Guangxuan and Lin, Ji and Seznec, Mickael and Wu, Hao and Demouth, Julien and Han, Song},
  booktitle={International conference on machine learning},
  pages={38087--38099},
  year={2023},
  organization={PMLR}
}

@article{tseng2024quip,
  title={Quip\#: Even better llm quantization with hadamard incoherence and lattice codebooks},
  author={Tseng, Albert and Chee, Jerry and Sun, Qingyao and Kuleshov, Volodymyr and De Sa, Christopher},
  journal={arXiv preprint arXiv:2402.04396},
  year={2024}
}

@article{dettmers2023spqr,
  title={Spqr: A sparse-quantized representation for near-lossless llm weight compression},
  author={Dettmers, Tim and Svirschevski, Ruslan and Egiazarian, Vage and Kuznedelev, Denis and Frantar, Elias and Ashkboos, Saleh and Borzunov, Alexander and Hoefler, Torsten and Alistarh, Dan},
  journal={arXiv preprint arXiv:2306.03078},
  year={2023}
}

@article{merity2016pointer,
  title={Pointer sentinel mixture models},
  author={Merity, Stephen and Xiong, Caiming and Bradbury, James and Socher, Richard},
  journal={arXiv preprint arXiv:1609.07843},
  year={2016}
}

@article{zhang2022opt,
  title={Opt: Open pre-trained transformer language models},
  author={Zhang, Susan and Roller, Stephen and Goyal, Naman and Artetxe, Mikel and Chen, Moya and Chen, Shuohui and Dewan, Christopher and Diab, Mona and Li, Xian and Lin, Xi Victoria and others},
  journal={arXiv preprint arXiv:2205.01068},
  year={2022}
}

@article{yang2025qwen3,
  title={Qwen3 technical report},
  author={Yang, An and Li, Anfeng and Yang, Baosong and Zhang, Beichen and Hui, Binyuan and Zheng, Bo and Yu, Bowen and Gao, Chang and Huang, Chengen and Lv, Chenxu and others},
  journal={arXiv preprint arXiv:2505.09388},
  year={2025}
}

@article{dubey2024llama,
  title={The llama 3 herd of models},
  author={Dubey, Abhimanyu and Jauhri, Abhinav and Pandey, Abhinav and Kadian, Abhishek and Al-Dahle, Ahmad and Letman, Aiesha and Mathur, Akhil and Schelten, Alan and Yang, Amy and Fan, Angela and others},
  journal={arXiv e-prints},
  pages={arXiv--2407},
  year={2024}
}

@software{autogptq,
  author = {PanQiWei and others},
  title = {AutoGPTQ: An easy-to-use LLMs quantization package with user-friendly APIs, based on GPTQ algorithm},
  year = {2023},
  url = {https://github.com/AutoGPTQ/AutoGPTQ},
  note = {GitHub repository}
}

@inproceedings{rosenthal2017semeval,
  title={SemEval-2017 task 4: Sentiment analysis in Twitter},
  author={Rosenthal, Sara and Farra, Noura and Nakov, Preslav},
  booktitle={Proceedings of the 11th international workshop on semantic evaluation (SemEval-2017)},
  pages={502--518},
  year={2017}
}

@inproceedings{mishra2019ocr,
  title={Ocr-vqa: Visual question answering by reading text in images},
  author={Mishra, Anand and Shekhar, Shashank and Singh, Ajeet Kumar and Chakraborty, Anirban},
  booktitle={2019 international conference on document analysis and recognition (ICDAR)},
  pages={947--952},
  year={2019},
  organization={IEEE}
}

@inproceedings{duan2024vlmevalkit,
  title={Vlmevalkit: An open-source toolkit for evaluating large multi-modality models},
  author={Duan, Haodong and Yang, Junming and Qiao, Yuxuan and Fang, Xinyu and Chen, Lin and Liu, Yuan and Dong, Xiaoyi and Zang, Yuhang and Zhang, Pan and Wang, Jiaqi and others},
  booktitle={Proceedings of the 32nd ACM International Conference on Multimedia},
  pages={11198--11201},
  year={2024}
}

\end{document}